\PassOptionsToPackage{unicode}{hyperref}
\PassOptionsToPackage{hyphens}{url}
\PassOptionsToPackage{dvipsnames,svgnames,x11names}{xcolor}
\documentclass[
  12pt]{article}

\usepackage{amsmath,amssymb}
\usepackage{iftex}
  \usepackage[T1]{fontenc}
  \usepackage[utf8]{inputenc}
  \usepackage{textcomp} % provide euro and other symbols
\usepackage{lmodern}
\IfFileExists{upquote.sty}{\usepackage{upquote}}{}
\IfFileExists{microtype.sty}{% use microtype if available
  \usepackage[]{microtype}
  \UseMicrotypeSet[protrusion]{basicmath} % disable protrusion for tt fonts
}{}
\makeatletter
\@ifundefined{KOMAClassName}{% if non-KOMA class
  \IfFileExists{parskip.sty}{%
    \usepackage{parskip}
  }{% else
    \setlength{\parindent}{0pt}
    \setlength{\parskip}{6pt plus 2pt minus 1pt}}
}{% if KOMA class
  \KOMAoptions{parskip=half}}
\makeatother
\usepackage{xcolor}
\makeatletter
\ifx\paragraph\undefined\else
  \let\oldparagraph\paragraph
  \renewcommand{\paragraph}{
    \@ifstar
      \xxxParagraphStar
      \xxxParagraphNoStar
  }
  \newcommand{\xxxParagraphStar}[1]{\oldparagraph*{#1}\mbox{}}
  \newcommand{\xxxParagraphNoStar}[1]{\oldparagraph{#1}\mbox{}}
\fi
\ifx\subparagraph\undefined\else
  \let\oldsubparagraph\subparagraph
  \renewcommand{\subparagraph}{
    \@ifstar
      \xxxSubParagraphStar
      \xxxSubParagraphNoStar
  }
  \newcommand{\xxxSubParagraphStar}[1]{\oldsubparagraph*{#1}\mbox{}}
  \newcommand{\xxxSubParagraphNoStar}[1]{\oldsubparagraph{#1}\mbox{}}
\fi
\makeatother

\usepackage{longtable,booktabs,array}
\usepackage{calc} % for calculating minipage widths
\usepackage{etoolbox}
\makeatletter
\patchcmd\longtable{\par}{\if@noskipsec\mbox{}\fi\par}{}{}
\makeatother
\IfFileExists{footnotehyper.sty}{\usepackage{footnotehyper}}{\usepackage{footnote}}
\makesavenoteenv{longtable}
\usepackage{graphicx}
\makeatletter
\def\maxwidth{\ifdim\Gin@nat@width>\linewidth\linewidth\else\Gin@nat@width\fi}
\def\maxheight{\ifdim\Gin@nat@height>\textheight\textheight\else\Gin@nat@height\fi}
\makeatother
\setkeys{Gin}{width=\maxwidth,height=\maxheight,keepaspectratio}
\makeatletter
\def\fps@figure{htbp}
\makeatother

\makeatletter
\@ifpackageloaded{caption}{}{\usepackage{caption}}
\AtBeginDocument{%
\ifdefined\contentsname
  \renewcommand*\contentsname{Table of contents}
\else
  \newcommand\contentsname{Table of contents}
\fi
\ifdefined\listfigurename
  \renewcommand*\listfigurename{List of Figures}
\else
  \newcommand\listfigurename{List of Figures}
\fi
\ifdefined\listtablename
  \renewcommand*\listtablename{List of Tables}
\else
  \newcommand\listtablename{List of Tables}
\fi
\ifdefined\figurename
  \renewcommand*\figurename{Figure}
\else
  \newcommand\figurename{Figure}
\fi
\ifdefined\tablename
  \renewcommand*\tablename{Table}
\else
  \newcommand\tablename{Table}
\fi
}
\@ifpackageloaded{float}{}{\usepackage{float}}
\floatstyle{ruled}
\@ifundefined{c@chapter}{\newfloat{codelisting}{h}{lop}}{\newfloat{codelisting}{h}{lop}[chapter]}
\floatname{codelisting}{Listing}

\makeatother
\makeatletter
\@ifpackageloaded{caption}{}{\usepackage{caption}}
\@ifpackageloaded{subcaption}{}{\usepackage{subcaption}}
\makeatother

\usepackage[]{natbib}
\usepackage{bookmark}

\IfFileExists{xurl.sty}{\usepackage{xurl}}{} % add URL line breaks if available
\hypersetup{
  pdftitle={Title},
  pdfauthor={Author 1; Author 2},
  pdfkeywords={3 to 6 keywords, that do not appear in the title},
  colorlinks=true,
  linkcolor={blue},
  filecolor={Maroon},
  citecolor={Blue},
  urlcolor={Blue},
  pdfcreator={LaTeX via pandoc}}

\newcommand{\anon}{1}

\usepackage{algorithm,algpseudocode}
\usepackage{amssymb}
\usepackage[export]{adjustbox}
\usepackage{multirow}

\usepackage{enumitem}
\usepackage{amsthm}
\usepackage{dsfont}

\newtheorem{definition}{Definition}

\newtheorem{proposition}{Proposition}
\newtheorem{remark}{Remark}

\newcommand{\new}[1]{\textcolor{blue!50!black}{#1}}
\renewcommand{\new}[1]{\textcolor{black}{#1}}

\newcommand{\eqdef}{\overset{\mathrm{def.}}{=}}
\newcommand{\Fc}{\mathcal{F}}
\newcommand{\Sc}{\mathcal{S}}
\newcommand{\Oc}{\mathcal{O}}
\newcommand{\Nc}{\mathcal{N}}
\newcommand{\Rc}{\mathcal{R}}

\newcommand{\e}{\mathbf{e}}

\newcommand{\Q}{\mathbf{Q}}
\newcommand{\U}{\mathbf{U}}
\newcommand{\I}{\mathbf{I}}
\renewcommand{\L}{\mathbf{L}}
\renewcommand{\v}{\mathbf{v}}
\newcommand{\x}{\mathbf{x}}
\newcommand{\y}{\mathbf{y}}
\newcommand{\Y}{\mathbf{Y}}
\newcommand{\X}{\mathbf{X}}
\newcommand{\z}{\mathbf{z}}
\renewcommand{\u}{\mathbf{u}}
\newcommand{\f}{\mathbf{f}}
\newcommand{\Z}{\mathbf{Z}}
\newcommand{\Zbold}{\mathbf{0}}

\newcommand{\bmu}{\boldsymbol{\mu}}

\newcommand{\bSig}{\boldsymbol{\Sigma}}

\newcommand{\bb}{\mathbf{b}}
\newcommand{\df}{\mathrm{DFT}}
\newcommand{\idf}{\mathrm{IDFT}}
\newcommand{\Rn}{\mathbb{R}^n}

\begin{document}

\def\spacingset#1{\renewcommand{\baselinestretch}%
{#1}\small\normalsize} \spacingset{1}

%%%%%%%%%%%%%%%%%%%%%%%%%%%%%%%%%%%%%%%%%%%%%%%%%%%%%%%%%%%%%%%%%%%%%%%%%%%%%%

\if1\anon
{
	%\title{\bf Lightweight sampling of Markov random fields}
    \title{\bf{A new class of Markov random fields enabling lightweight sampling}}
	\author{Jean-Baptiste Courbot\\
		IRIMAS UR 7499, Université de Haute-Alsace, Mulhouse, France\\
		and \\
		Hugo Gangloff \\
		Université Paris-Saclay, AgroParisTech, INRAE UMR MIA Paris-Saclay, France \\
		and \\
		Bruno Colicchio\\
		IRIMAS UR 7499, Université de Haute-Alsace, Mulhouse, France
		}
	\maketitle
} \fi

\if0\anon
{
	\bigskip
	\bigskip
	\bigskip
	\begin{center}
		{\LARGE\bf A new class of Markov random fields enabling lightweight sampling}
	\end{center}
	\medskip
} \fi

%
%\if1\anon
%{
%  \title{\bf Lightweight sampling of Markov random fields}
%  \author{Author 1\thanks{
%    The authors gratefully acknowledge \textit{please remember to list all relevant funding sources in the version that gives all author information}}\hspace{.2cm}\\
%    Department of YYY, University of XXX\\
%    and \\
%    Author 2 \\
%    Department of ZZZ, University of WWW}
%  \maketitle
%} \fi
%
%\if0\anon
%{
%  \bigskip
%  \bigskip
%  \bigskip
%  \begin{center}
%    {\LARGE\bf Title}
%\end{center}
%  \medskip
%} \fi

\bigskip
\begin{abstract}
This work addresses the problem of efficient sampling of Markov random fields (MRF).
The sampling of Potts or Ising MRF is most often based on Gibbs sampling, and is thus computationally expensive.
We consider in this work how to circumvent this bottleneck through a link with Gaussian Markov Random fields. The latter can be sampled in several cost-effective ways, and we introduce a mapping from real-valued GMRF to discrete-valued MRF.
The resulting new class of MRF benefits from a few theoretical properties that validate the new model. Numerical results show the drastic performance gain in terms of computational efficiency, as we sample at least 35x faster than Gibbs sampling using at least 37x less energy, all the while exhibiting empirical properties close to classical MRFs.
\end{abstract}

\noindent%
\vfill
%\todo{3 to 6 keywords, that do not appear in the title}

%\todo{Penser à remettre le spacing à 1.8 !}

\newpage
%\spacingset{1.8} % DON'T change the spacing!

\section{Introduction}

\new{In this paper, we consider two  classes of random fields: Markov Random fields (MRFs) and Gaussian Markov Random Fields (GMRFs). This introduction summarizes their main properties as well as procedures to sample from these processes.}

\subsection{Markov random fields}

% Key points : on a lattice, discrete MRF, continuous GRF, Markovian GRF

%\CommentHugo{Emphasize that it is for discrete MRF ? Or I don't know to what extent is derives easily from what I wrote in B. GRF. I think this transition from continuous to discrete is critical is the work.}

%\CommentHugo{Thinking about it: we want to link discrete MRF on graph with continuous GRF (MRF) on continuous space. Here the Hammersly Clifford only holds in graph (finite set indexation setting right?). We need some biblio for all this. Are their some spectral measure equivalent for discrete MRF ? Can we look for discretization of GRF ? }

%\CommentHugo{No, Hammersly Clifford theorem is **not** only for discrete value X!}

\subsubsection{Definition} In the context of image processing, Markov Random Fields (MRFs) are a class of statistical processes that are often used to represent latent (or hidden) processes of interest, typically in segmentation tasks.
%discretely valued latent processes in image processing, typically in segmentation tasks.
Let $\Sc$ be the set of $n$ sites in an image, and let us denote $\X = \{X_s\}_{s\in\Sc}$. Denoting $\X$ a random vector and $\x$ its realization, we will also note, for brevity, the distributions $p(\X = \x)$ as $p(\x)$ when there is no ambiguity.

\new{The Markov property for $\X$ is the following.
	$\X$ is a MRF if and only if there exists a neighboring system $N$ such that }$\forall s$:
\begin{equation}
	p(x_s | \x_{\Sc \setminus s}) =  p(x_s | \x_{N_s}). \label{eq:markov}
\end{equation}
\new{The neighboring system can be chosen at discretion, but usually refers to 4- or 8-sets of pixels adjacent to the location $s$.} 
Thanks to the Hammersley-Clifford theorem~\citep{clifford1971markov}, and considering only pairwise site interactions,  the density of the process can be written as:
\begin{equation}
	p(\x) = \frac{1}{\gamma}\exp\left( - \sum_{s\in\Sc}\sum_{s' \in N_s} \psi(x_s,x_{s'}) \right),
\end{equation}
where $\psi$ is a pairwise potential and $\gamma$ is the partition function. Typically, one can choose for $\beta> 0$:
\begin{equation}
\begin{array}{crcll}
    & \psi(x_s,x_{s'})& = &\beta \delta_{\{x_s=x_{s'}\}}& \text{(Potts potential)},  \\
   \text{or} & \psi(x_s,x_{s'}) &=& \beta (1- \delta_{\{x_s=x_{s'}\}})& \text{(Ising potential)},
\end{array}\label{eq:ising_potts}
\end{equation}
with $\delta$ the Kronecker function. Later, the corresponding fields will be referred to as Potts or Ising MRFs, respectively.

%a pairwise interaction potential is $$, with $\beta>0$ \new{and $\delta$ the Kronecker function. This potential is usually referred as a Potts potential}, and such field will be referred to as Potts MRF in the following.

\subsubsection{Sampling} 
\new{Computing $\gamma$ is in the general case intractable as it requires to sum over all possible configurations of the $n$-valued random vector $\x$.
	Thus,} realizations $\X = \x$ are approximated  through iterative sampling techniques. This is mainly performed from Gibbs sampling as proposed by~\citep{geman1984stochastic}.
Because this sampling has a high computational cost, various efforts have been made in the literature to approximate $\gamma$~\citep{giovannelli2007unsupervised}, to find other models~\citep{laferte2000discrete}, or to improve updates within the Metropolis-Hastings sampler~\citep{grathwohl2021oops}.
Gibbs sampling nevertheless remains overall the main technique for sampling MRFs. 
%In this paper, we keep Gibbs-based sampling of Markov fields as our main comparison point, as most of the existing works are based on this technique.

As Gibbs sampling is a MCMC algorithm, its computational cost is mainly dependent on the expected time it takes to approximate the target distribution. This depends on the mixing rate of the underlying Markov chain, which in turn depends on the eigenvalues of its transition matrix.
For instance, \cite{frigessi1997computational} discuss the convergence time $t^*$, that is, the number of individual site update the Gibbs sample requires to reach an arbitrary precision, depending on the number of sites $n$. In their Theorem 1,
%For instance, in \cite[Theorem 1]{frigessi1997computational}, 
the authors show, 
%\new{for $n$ sites}, a convergence time $t^*$ 
in the Ising case that
$t^* \leq \Oc(n\log (n))$ 
%$\Oc(n\log (n))$ 
or 
$t^*<\Oc(n^2\log (n))$ ;
when there is no phase transition, \textit{i.e.} for small values of the Ising parameter $\beta$.

\new{Such values however imply that the resulting realization has no large homogeneous area, and thus these results are no relevant in } most of image-processing based situations. 
For large values of $\beta$ and in dimension 2, a higher bound is found as $t^* \leq \Oc( \exp (\beta C \sqrt{n}))$, with $C$ a constant. %\todo{[voir/ introduire les notations pour ce paragraphe]}
%\new{Note that}
%\new{}
Overall, works dealing with the convergence of Potts MRF sampling are quite seldom, and the results are also obtained for values of the Potts parameter that also makes them irrelevant for image processing tasks (see, \textit{e.g.},\citep{gheissari2016mixing,mossel2013exact}).

%\new{Note on the relevance for image processing of the works from statistical mechanics}

More recent work proposes using auxiliary Gaussian variables in an MCMC scheme in order to assist in sampling Ising~\citep{martens2010parallelizable} and Potts~\citep{margossian2021simulating} distributions, enabling in the latter case a computational complexity of $\Oc(n^3 + m n^2 K)$, $K$ being the number of classes and $m$ the number of Gibbs iterations.  %in the Ising augmentation to sample jointly with the target Ising
Other propositions involve the reformulation of Gibbs sampling over relevant sub-lattices, leading to \textit{chromatic}~\citep{gonzalez2011parallel} or \textit{conclique}~\citep{kaplan2020simulating} Gibbs sampling. 
The computational cost of a chromatic Gibbs sampler is $\Oc ( \frac{n}{q} + c)$ with $c$ colors (\textit{i.e.}, $c$ independent sub-lattices) and $q$ processors over $n$ sites. However, the works reported in this paragraph address the computational cost for a single Gibbs path over the grid, and thus the statistical convergence cost overhead must also be accounted for.
\new{To our knowledge, the chromatic Gibbs sampler is, to date, the most relevant sampler for Ising/Potts MRF when dealing with large images. It is reported in Appendix~\ref{ap:sample_mrf} and will serve as a reference for comparison in the numerical section.}

\subsection{Gaussian Markov random fields}

\subsubsection{Definition} Gaussian random fields (GRFs) are another class of statistical processes that are commonly used to model spatial dependencies between real-valued variables. In the case of image processing, GRFs may also be indexed by a lattice.
Let us define $\Z$ as a GRF on the lattice $\Sc$ iff $\Z \sim \Nc(\bmu, \bSig)$, where $\bmu \in \mathbb{R}^S$ and $\bSig \in \mathbb{R}^{S\times S}$ are the mean vector and the covariance matrix, respectively.

Upon known conditions on $\bSig$~\cite[p. 120]{rozanov1982markov}, GRFs are also Markovian, and are then referred to as Gaussian Markov random fields (GMRFs). These conditions imply that the precision matrix $\Q = \bSig^{-1}$ is sparse, in other words, that conditionally to any given site $s\in \Sc$, the correlation spans a limited neighborhood, yielding a Markov property similar to~\eqref{eq:markov}.

\subsubsection{Sampling} \label{subsec:sampling} 
The algorithmic complexity of GMRF sampling relies on the assumptions one is willing to make on $\bSig$. 
Following~\citep{rue2005gaussian}, the simplest option is to rely on a Cholesky decomposition of $\bSig$. However, this is rather ineffective as $n$ grows, \new{as its computational complexity is $\Oc (n^3)$}.

% as its cost is $\mathcal{O}(n^3)$.
\textit{Fourier sampling.} On the other hand, assuming that $\bSig$ is circulant enables computations based on discrete Fourier sampling. 
Thus, under this assumption, there is no need to compute or store neither $\Q$ nor $\bSig$ as their bases are sufficient to perform computations. % the Fourier domain.
However, sampling under this assumption implies that $\Sc$ is joined at the sides, making it a torus, and that $\Z$ is stationary on $\Sc$, \textit{i.e.}, that it locally behaves similarly regardless of the site $s$.

%\todo{equation Fourier sampling}

\textit{Spectral sampling.} More recent works~\citep{allard2020simulating} leveraged the torus constraint by introducing a \textit{spectral} sampling method.   %but requires sampling from gamma distribution which is slower than the normal sampling when using FFt on torus.
The only requirement  here is that the covariance function belongs to an extended Gneiting class of covariances, which includes in particular Matérn correlation functions.

Fourier and spectral sampling of GMRFs are both computationally efficient and will be considered later on. Details of their implementations are reported in Appendix~\ref{ap:sample_gmrf}.

\subsection{This paper}

\new{This paper introduces a novel class of Markov random fields, designed such that samples can be obtained through GMRF sampling.
	In Section~\ref{sec:model} we describe the processes involved and their key properties, as well as the sampling procedure.
	Section~\ref{sec:numeric} proposes then a numerical study, focusing on the sampling time and energy consumption, and on the empirical properties of sampled fields. 
	Then, Section~\ref{sec:discussion} wraps up the results and highlights some relations to other works.
	Finally, we gather in the Appendix the main algorithms mentioned in the paper, as well as additional numerical results.
\if1\anon
{ Note also that a preliminary version of this work was presented in~\citep{courbot25gums}.
} \fi
   }

\new{We also use a few hypotheses and notational conventions through this paper, that we summarize here.
	We assume all random processes are indexed by a finite grid $\Sc$. 
	Discrete-valued MRFs will take their values $\omega$ in the $K$-class set $\Omega \eqdef \{ \omega_0, \omega_1, \ldots, \omega_{K-1} \} \subseteq \mathbb{N}$.
	Bold notation will refer to vectors, matrices, or random processes, while non-bold notation will refer to scalar values or random variables.
	Besides, $\Zbold_N$ will refer to a zero-valued vector of size $N$ while $\I_N$ will refer to a square identity matrix of size $N\times N$, and Euclidean distances will be denoted $\| \cdot \|$.
}
%Functions and operators will be denoted 

% the following notational convention:
% \begin{itemize}
	%     \item bold and caps
	%     \item $\Sc$ and $s$ and $|\Sc| = n$
	%     \item 
	% \end{itemize}
%In this paper, we introduce a class of Markov fields that can be obtained from GMRF samples, and thus with the same computational cost.
%reaching thus for which sampling is made at the same speed as GMRF, reaching $\Oc(N\log(N))$ in the best scenario. 
%Section~\ref{sec:model} introduces the model, while Section~\ref{sec:numeric} demonstrates the results in practice.

\section{Gaussian Unit simplex Markov random fields}
\label{sec:model}

% Peu utile :
%This section introduces a specific class of random fields and shows how it can be used to sample a discrete random field in a computationally efficient fashion. %in a 
%This section defines the random field we introduces and presents some of its properties.
%The 

\subsection{Definitions}

\begin{figure}
	\centering
	\,\hfill
	\subfloat[GMRF realization $\z \in \mathbb{R}^n$. ]{\includegraphics[frame,width=0.25\linewidth]{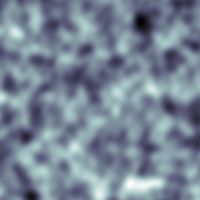}}\hfill 
	\subfloat[Discretized field $\f \in \{0,1\}^{n}$ such that $\forall s\in \Sc,  f_s = \mathds{1}_{\{z_s > 0\}}$.]{\includegraphics[frame,width=0.25\linewidth]{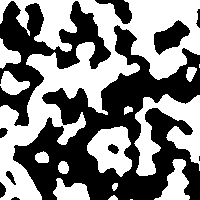}}\hfill
	\subfloat[Realization of a Potts MRF ($K=1$).]{\includegraphics[frame,width=0.25\linewidth]{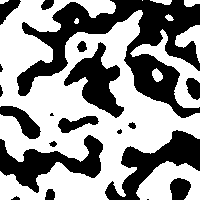}}\hfill\,
	\caption{Original intuition motivating this work: thresholding a GMRF realization (a) yields a discrete random field (b) that behaves similarly to a discrete MRF realization (c).}
	\label{fig:intuition}
\end{figure}

\new{The intuition underlying the link between continuous GMRF and discrete MRFs is the following: a thresholded version of a GMRF yields a two-class field that is visually similar to a binary MRF (see Fig.~\ref{fig:intuition}). In this section, we propose an expansion beyond the intuitive $K=2$ classes case and using a continuous function of a GMRF $\Z$ to perform the mapping towards the discrete case. 
	The expansion in more than 2 dimensions requires handling more than one GMRF realization and splitting the space they take value in. To do so, we use the following definition of unit simplex.} %to be able to partition a space $\mathbb{R}$

%\todo{add a figure for the binary case}

%This section describes the random field we propose and some of its properties, based on the following definition of unit simplex.

\begin{definition}[Unit simplex]
	\label{def:unit_simplex}
	A unit P-simplex is a regular simplex belonging in $\mathbb{R}^P$, whose $P+1$ vertices lie on a unit sphere. From~\citep{anderson2021coordinate} we take its vertices $\v \in \mathbb{R}^P$ as:% $\v_{P+1} = \frac{-1}{\sqrt{P}}\mathbf{1}$ and :
	\begin{equation}
		\v_j = \sqrt{\frac{P+1}{P}} \e_j - \frac{\sqrt{P+1} - 1}{P\sqrt{P}}\mathbf{1}~ \forall 1\leq j \leq P \text{ and } \v_{P+1} = \frac{-1}{\sqrt{P}}\mathbf{1}
	\end{equation}
	with $\mathbf{1} \in \mathbb{R}^P$ a vector of ones and $\e_j\in \mathbb{R}^P$ the $j-$th basis vector. 
	
	Let us denote $\mathbf{u}_P = \{\v_1, \ldots, \v_{P+1}\}$ the set of vertices of the unit $P$-simplex. Note that these are coordinate-invariant, so any permutation also lies on the unit sphere.
\end{definition}

\new{The second ingredient towards a generalized thresholding lies in the definition of a multivariate GMRF, which is defined as a GMRF with vector values on each $s\in \Sc$. Throughout this section, we will handle the multivariate GMRF $\Z$ such that for a given number of classes $K$:
	\begin{equation}
		\Z \sim \Nc(\bmu, \bSig) \text{ with mean } \bmu \in \mathbb{R}^{n(K-1)} \text{ and covariance matrix } \bSig \in \mathbb{R}^{n(K-1) \times n(K-1)}
		\label{eq:m-gmrf}
	\end{equation}
	Each random variable within $\Z$ is indexed both in terms of location on the grid $\Sc$, and in terms of \emph{component} among the $K-1$ values given at each $s$. Later in the paper, we will also refer to {components} of $\Z$, that is, the $K-1$ fields $\Z_k$ taking values in $\mathbb{R}^n$.  We are now equipped to define a mapping of $\Z=\{\Z_k\}_{k=0}^{K-2}$ with respect to vertices of a unit simplex.}

\begin{figure}[h]

	~~~~~\subfloat[$\Z=\z$: first component $\z_0$]{~\includegraphics[height=0.25\linewidth]{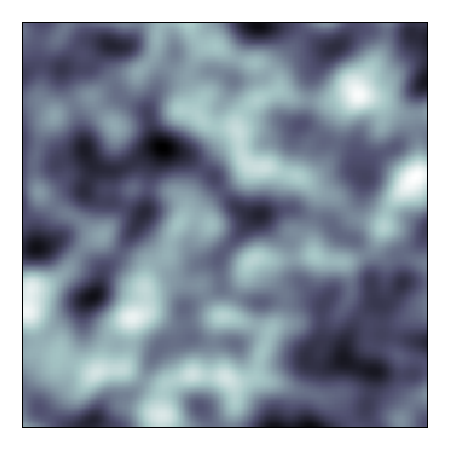}}~~~~~
	\subfloat[$\Z=\z$: second component $\z_1$.]{\includegraphics[height=0.25\linewidth]{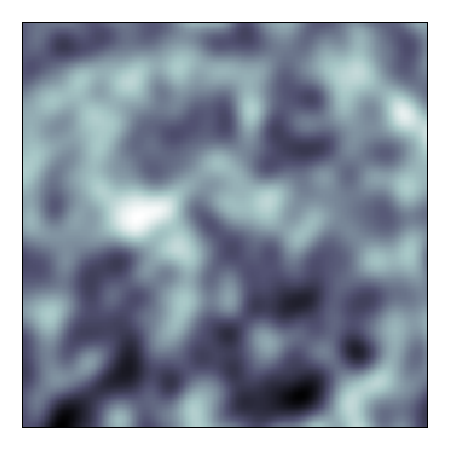}}~~~~~
	\subfloat[$\Z=\z$ projected in $\mathbb{R}^2$, and locations of the unit simplex vertices.]{\includegraphics[height=0.25\linewidth]{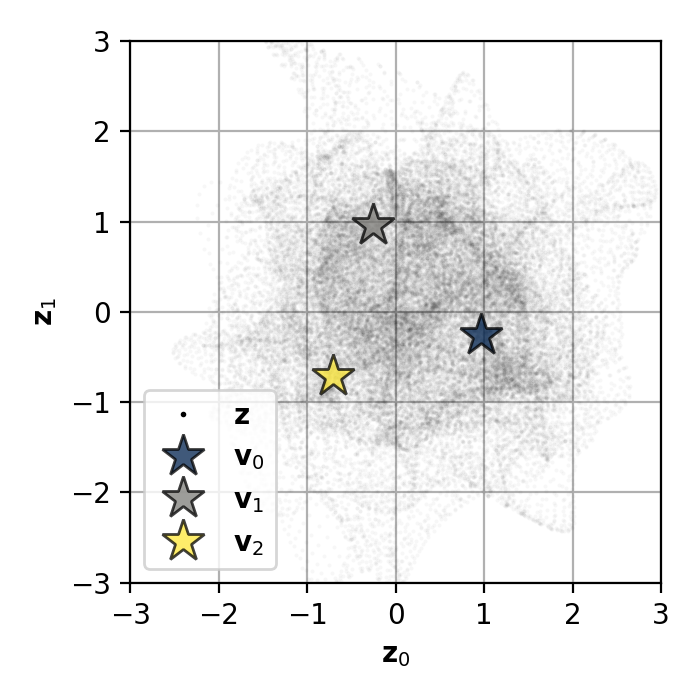}~~~~}\\
	
	~~~~~\subfloat[$\pi_0^c (\z)$.]{\includegraphics[height=0.25\linewidth]{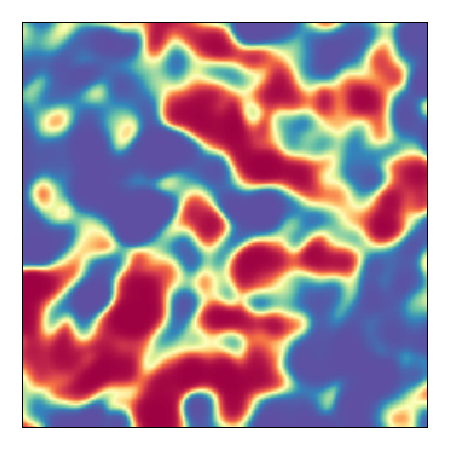}}~~~~~
	\subfloat[$\pi_1^c(\z)$.]{\includegraphics[height=0.25\linewidth]{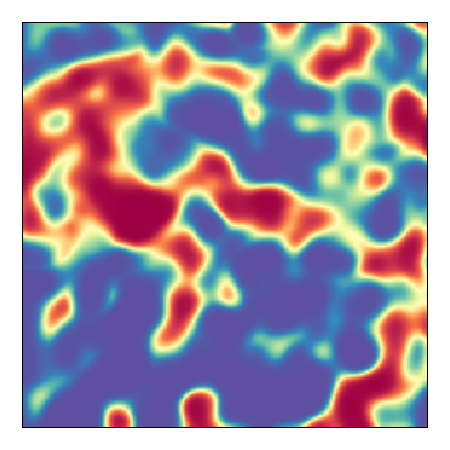}}~~~~~
	\subfloat[$\pi_2^c(\z)$.]{\includegraphics[height=0.25\linewidth]{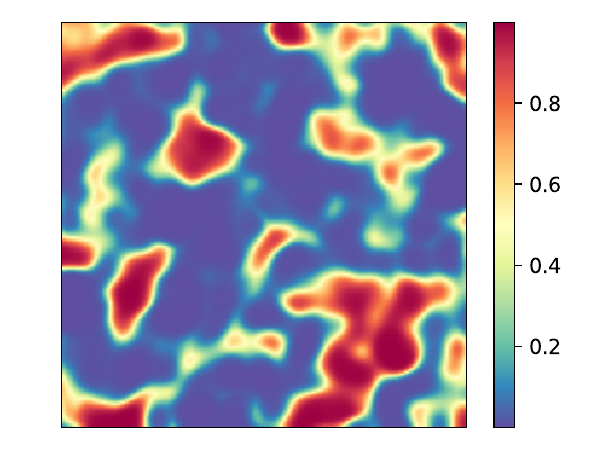}}\\
	\caption{Depiction of a realization of $\Z$ in the $K=3$ case, together with the mapping $\pi_k^c$, with $c=1$. \label{fig:pi_i}}
\end{figure}

\begin{definition}[Mapping with respect to unit simplex]
	\label{def-pi}
	\new{Let $c > 0$.
		We design $\pi_i^c:\mathbb{R}^{K-1}\mapsto [0,1]$ to be a measure of the distance between $\Z$ and the $i-$th vertice $\v_i$ of $\mathbf{u}_{K-1}$, such that $\forall s \in \Sc$:
		\begin{equation}
			\pi_i^c(\Z_s) = \frac{\exp(-c^{-2}\|\Z_s - \v_i\|^2)}{\sum_{k=0}^{K-1} \exp(-c^{-2}\|\Z_s - \v_k\|^2)}
			\label{eq:pi}
		\end{equation}
		with $c > 0$ and $\v_k, \v_i \in \mathbf{u}_{K-1}$ unit simplex vertices. %, and $\omega_i \in \Omega \subset \mathbb{N}$.
     %$\pi_i^c$ thus measure the distance of $\z_s$ to $\v_i$, seen from a Gaussian kernel, and normalized. 
        }
\end{definition}
\new{An illustration of the application of $\pi_i^c$ to a realization $\Z=\z$ is given in Fig.~\ref{fig:pi_i}.
}

\begin{figure}[h]
	\centering 
	\subfloat[$c=1$]{\includegraphics[width=0.25\linewidth]{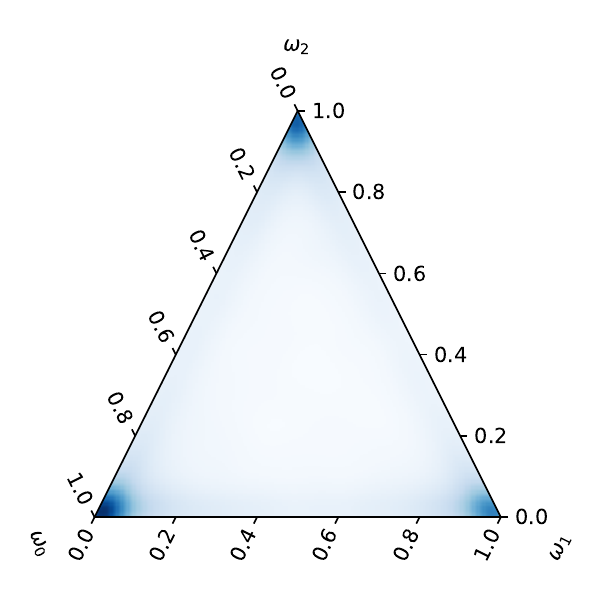}}
	\subfloat[$c=0.5$]{\includegraphics[width=0.25\linewidth]{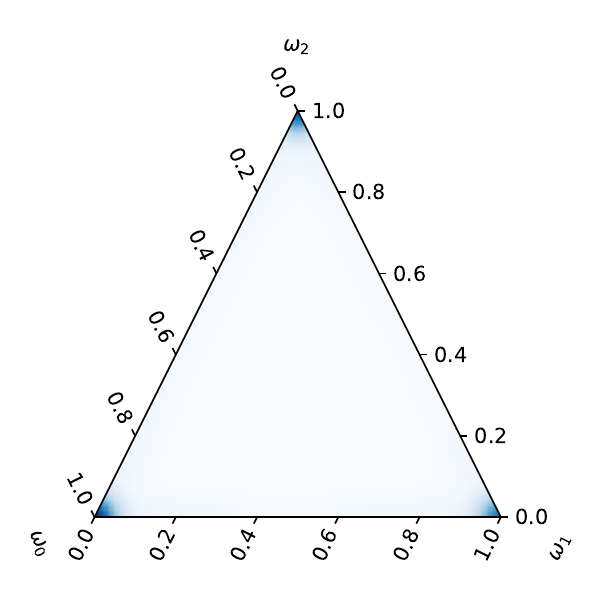}}
	\subfloat[$c=0.25$]{\includegraphics[width=0.25\linewidth]{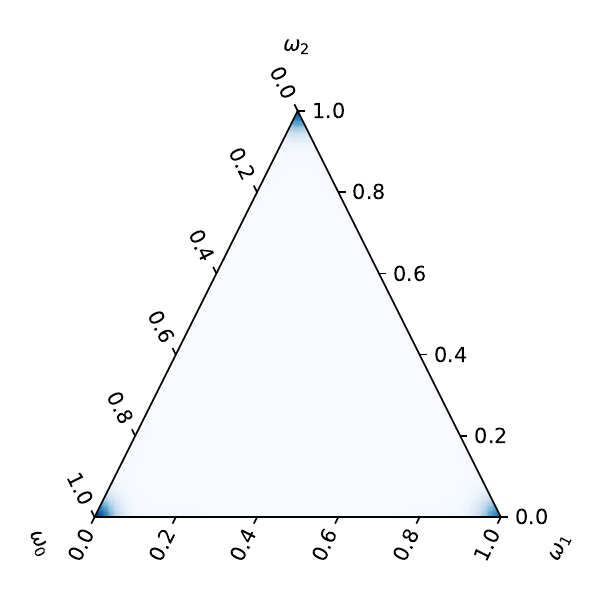}}
	\caption{Kernel density estimation obtained from the distribution of the triplet $\big(\pi_0^c(z_s),\pi_1^c(z_s),\pi_2^c(z_s)\big)$ in the probability simplex whose vertices locates classes in $\Omega=\{\omega_0,\omega_1,\omega_2\}$, over all $s\in \Sc$ for a given realization $\Z=\z$. As $c \rightarrow \infty$, the distribution tends towards Dirac masses located at the vertices of the probability simplex.}
	\label{fig:prob_simplex}
\end{figure}

\begin{remark}\new{Besides, as $\sum_{i=0}^{K-1}\pi_i^c (\Z_s) = 1$, the set $\{\pi_i^c (\Z_s)\}_{0\leq i\leq K-1}$ can be seen as a set of probabilities and represented on the \emph{probability} simplex. Figure~\ref{fig:prob_simplex} depicts this mapping for several values of $c$ in the $K=3$ case.
		This mapping highlights the similarity of the mapping~\eqref{eq:pi} with other distributions, namely the logit-normal multivariate distribution, and the Concrete or Gumbel softmax distributions~\citep{maddison2017concrete,jang2017categorical}. 
		While its behavior at the limit of $c \rightarrow \infty$ is similar to those distributions, the distribution induced by~\eqref{eq:pi} is however different:
		\begin{itemize}
			\item it can be seen as a softmax of \emph{distances} of Gaussian variables of the vertices of the simplex.
			\item it yields spatial correlations as $\Z$ is a GMRF.
			\item our purpose here is sampling discrete, spatially correlated variables.
            %, rather than relaxing a discrete-valued function to allow gradient computations to propagate through neural networks.
	\end{itemize}}
\end{remark}
% voir aussi
% https://arxiv.org/pdf/1611.01144
% https://arxiv.org/pdf/1611.00712

% \begin{remark}
	
	% \end{remark}

\begin{definition}[Gaussian Unit-simplex Markov random field (GUM)]
	\label{def:gum}
	Let $\mathbf{u}_{K-1}$ be a unit $(K-1)$-simplex (Definition~\ref{def:unit_simplex}), $\Z$ a multivariate GMRF (Eq.~\eqref{eq:m-gmrf}), the set of classes $\Omega = \{\omega_0, \ldots, \omega_{K-1} \} \subseteq \mathbb{N}$, and $c > 0$.
	%and $\Z \sim \Nc(\mathbf{0}, \bSig)$ a GMRF taking values in $\mathbb{R}^{n(K-1)}$, such that $\Z = \{\Z_s \}_{s\in\Sc}$ and $\Z_s$ takes values in $\mathbb{R}^{K-1}$.
	
	We define the mapping $\phi_{K,c}\colon \mathbb{R}^{n(K-1)} \mapsto \mathbb{R}^n$ such that:
	\begin{align}
		\phi_{K,c}(\Z) &= \sum_{i=0}^{K-1} \omega_i \pi^c_{i}(\Z) \label{eq:phi}
	\end{align}
	$\phi_{K,c}(\Z)$ is named a \emph{GUM random field}.
\end{definition}
This transformation can be seen as a class-weighted measure of distance (seen from a Gaussian distribution of standard deviation $c$) between the points in $\Z=\z$ and the vertices of the unit simplex.

\begin{remark}
    Note that for this mapping to be well-defined, we need to represent classes with a set $\Omega \subseteq \mathbb{\mathbb{R}^+}$. We set $\Omega \subseteq \mathbb{N}$ for this purpose. We consider this is not limiting, as any discrete set can be indexed in $\mathbb{N}$ and thus mapped to positive integers.
\end{remark}

\begin{figure}[t]
	\flushleft 
	%\subfloat[Realization $\z$: first component]{~\includegraphics[height=3.75cm]{fig_1/z1_no_cb2.png}}
	%\subfloat[Realization $\z$: second component]{\includegraphics[height=3.75cm]{fig_1/z22.png}} 
	\subfloat[$\phi_{K,c}(\z)$ with $c = 1 $.]{\includegraphics[height=0.25\linewidth]{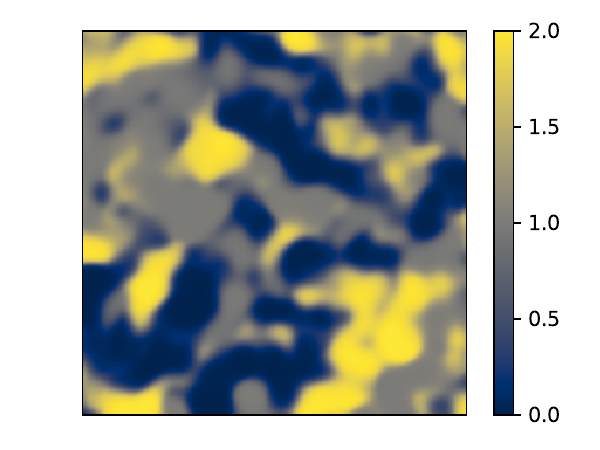}}~%
	\subfloat[$\phi_{K,c}(\z)$ with $c = 0.5$.]{\includegraphics[height=0.25\linewidth]{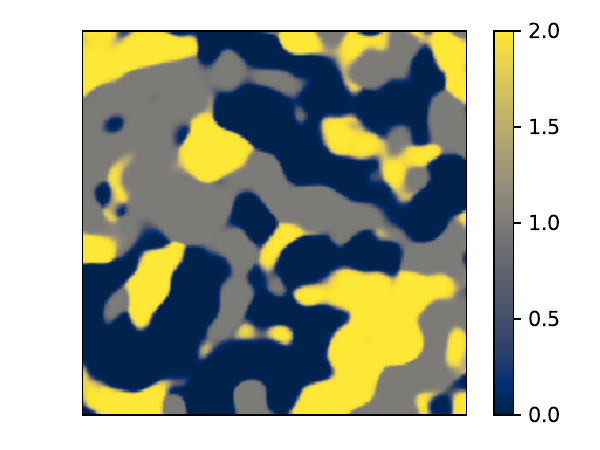}}
	\subfloat[$\phi_{K,c}(\z)$with $c = 0.25$.]{~\includegraphics[height=0.25\linewidth]{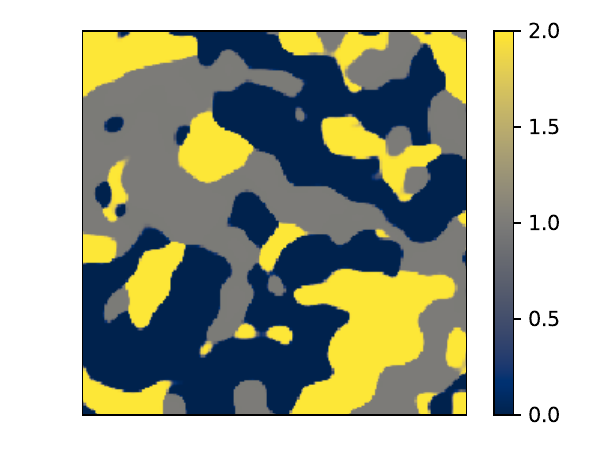}}\\%
	\subfloat[DGUM of $\z$.]{\includegraphics[height=0.25\linewidth]{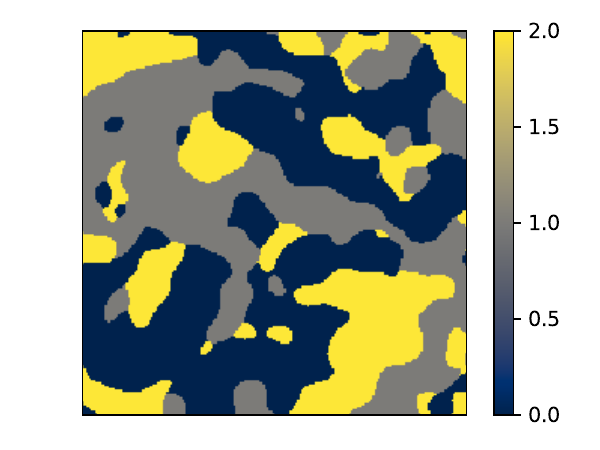}~~~}
	%\subfloat[Realization $\z$ where each site value is located in $\mathbb{R}^{K-1}$, and locations of the unit simplex vertices.]{\includegraphics[height=4.25cm]{fig_1/z1_z2_2D2.png}}~%
	\subfloat[$\Z=\z$ colored according to the classes in $\x$. \label{fig:2d_gum}]{\includegraphics[height=0.25\linewidth]{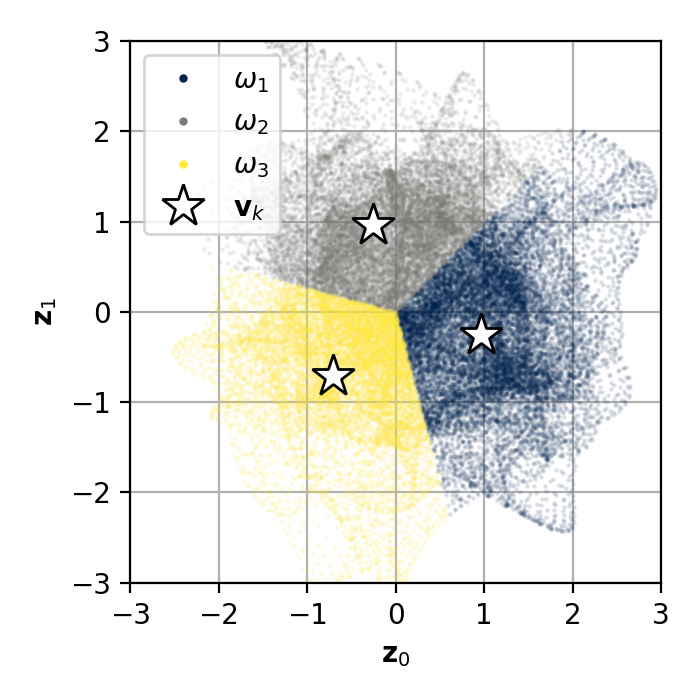}}
	\caption{Illustration of the DGUM sampling for $K=3$ classes. The sampling is performed from $\Z=\z$ depicted in Fig.~\ref{fig:pi_i}. (a)-(c) depict the GUM realization~\eqref{eq:phi}, and (d) depicts its limit DGUM realization $\x$~\eqref{eq:dgum}. (e) depicts the classes in $\x$ for the values of $\z$ in $\mathbb{R}^2$. \label{fig:gum_formation}}
\end{figure}

We finally introduce a limiting process that will yield a discretization of $\phi_{K,c}(\Z)$.
\begin{definition}[Discrete GUM]
	\label{def:digum}
	Let $\Z$ be a GMRF. 
	From Definition~\ref{def:gum}, we have:% that when $c\rightarrow 0$,  :
	\begin{equation}
		\phi_{K,c}(\Z) \underset{c\rightarrow 0}{\longrightarrow} \sum_{i=0}^{K-1} \omega_i \delta_{\big[\|\Z - \v_i\| \leq \|\Z - \v_k\|, ~\forall \v_k \in \mathbf{U}_{K-1}\big]}
	\end{equation}
	Denoting $ \underset{c\rightarrow 0}{\lim}\,\phi_{K,c}(\Z) =\phi_K(\Z) =\X = \{X_s\}_{s\in\Sc} $, we have $\forall s \in \Sc$:
	\begin{equation}
		X_s = \omega_{k^*} \text{ with $k^*$ chosen such that } \v_{k^*} = \underset{\v \in \U_{K-1}}{\arg \min} \,\|\Z_s - \v  \|
		\label{eq:dgum}
		%  \hat{x}^{\text{MAP}}_s = \omega_{k^*} \text{ with $k^*$ chosen such that } \v_{k^*} = \underset{\v \in \U_{K-1}}{\arg \min} \,\|\z_s - \v  \| \label{eq:map}
	\end{equation}
	This discrete limit process will be referred to as a \emph{Discretized GUM} or \emph{DGUM}.
\end{definition}
In other words, the mapping $\Z \mapsto \underset{c\rightarrow 0}{\lim}\,\phi_{K,c}(\Z)$ indicates pointwise from which vertices among the unit $(K-1)$-simplex vertices each $\z_s$ is the closest. 
The  process linking a realization of a GMRF $\Z=\z$ to its DGUM is reported in Fig.~\ref{fig:gum_formation}.

% We assume here and in the following some hypothesis regarding the GMRF $\Z$ we use to design the DGUM from Definition~\ref{def:digum}, in order to simplify the discussion:
% \begin{itemize}
	% 	\item for a given site $s$, the $K-1$ components of $\Z_s$ are independent, thus, $\Z$ can be seen as $K-1$ independent GMRF.
	% 	\item  We also assume they all share the same covariance matrix. % $\bSig$
	% \end{itemize}

\subsection{Properties}

In this section, we state and prove a few properties of the GUM and DGUM processes defined in the previous section in order to better understand their theoretical behavior. % the processes we consider. % our understanding of the processes. 
%We first prove that GUM and DGUM are also Markov fields, before moving to the s

\begin{proposition}[Markovianity of $\phi_{K,c}(\Z)$]
	\label{prop:gum}
	Let $\Z$ be a GMRF. Then, $\phi_{K,c}(\Z)$ is a Markov random field.
\end{proposition}
\begin{proof}
	Let us denote $\Fc(A)$ the $\sigma-$algebra induced by a random variable $A$.
	Rephrasing Equation~\eqref{eq:markov}, $\Z$ is Markovian if and only if there exists a neighborhood system $N$ such that for any $s\in\Sc$, $\Fc(\Z_{N_s})$ is splitting $\Fc(\Z_s)$ and $\Fc(\Z_{\Sc \setminus s, N_s})$ (see~\cite[Chap. 2]{rozanov1982markov}). 
	
	\new{Besides, $\phi_{K,c}$ being continuous, it is a  Borel function.}
	Then Doob-Dynkin lemma~\citep{kallenberg1997foundations} states that its application on random variables contracts the sigma-algebra, \new{that is: $\Fc({\phi_{K,c}(\Z_s)}) \subseteq \Fc({\Z_s})$, $\Fc({\phi_{K,c}(\Z_{N_s})}) \subseteq \Fc({\Z_{N_s}})$ and $\Fc({\phi_{K,c}(\Sc \setminus s, N_s)}) \subseteq \Fc({\Sc \setminus s, N_s})$. }
	Hence, $\Fc(\phi_{K,c}(\Z_{N_s}))$ is also splitting $\Fc(\phi_{K,c}(\Z_s))$ and $\Fc(\phi_{K,c}(\Z_{\Sc \setminus s, N_s}))$. Thus, $\phi_{K,c}(\Z)$ is also Markovian.
	
\end{proof}

\begin{proposition}[\new{Markovianity of $\X$}] \new{Let $\Z$ be a GMRF. Then, the DGUM $\phi_K(\Z)$ from Definition~\ref{def:digum} is also Markovian.}
\end{proposition}
\begin{proof}
	\new{In the proof of Proposition~\ref{prop:gum}, we saw that $\phi_{K,c}$ being a Borel function, it preserves the Markov property assumed in $\Z$.
		Let us index $c$ by $u\in\mathbb{N}$ such that $\{c_u\}_{u \in \mathbb{N}}$ decreases and $\underset{u\rightarrow \infty}{\lim} c_u = 0$.
		Then, $\X = \underset{c\rightarrow 0}{\lim}\,\phi_{K,c}(\Z) = \underset{u\rightarrow \infty}{\lim} \,\phi_{K,c_u}(\Z)$, so that $\phi_K$ is the pointwise limit of $\phi_{K,c_u}$. As the latter is a Borel function, then $\phi_K$ is also a Borel function. 
		Thus, $\phi_K(\Z)$ is also Markovian.}
\end{proof}
% Source livre :  http://tomlr.free.fr/Math%E9matiques/Math%20Complete/Probability%20and%20statistics/Foundations%20of%20Modern%20Probability%20-%20Olav%20Kallenberg.pdf (voir)
% https://mathoverflow.net/questions/267788/are-bochner-measurablity-and-borel-measurability-compatible

\begin{remark}
	The Markovianity of $\X$, provided that $\Z$ is Markovian, implies that $\X$ and $\Z$ share the same neighborhood structure. In practice, the covariance matrix of $\Z$ might dampen the correlation; thus the observed $\X$ may have a smaller apparent neighborhood dependency than the one described from $\Z$.

\end{remark}
We now specify further the conditions to reach class balance in DGUM realization, for which $\Z$ is required to be centered and isotropic.

\begin{definition}[Balanced and unbalanced GUMs]
	\label{def:balance}
	\new{
		Let us denote $\bmu \in \mathbb{R}^{K-1}$ the vector of scalar means of the $(K-1)$ random fields composing $\Z$. 	When $\bmu =  \mathbf{0}_{K-1}$, we denote $\Z$ as a \emph{balanced} GUM. Otherwise, it will be referred to as \emph{unbalanced}.
		We similarly denote balanced and unbalanced DGUM from Definition~\ref{def:digum}.}
\end{definition}

\begin{definition}[Isotropic and anisotropic GUMs]
	\label{def:isotropic}
	%We further refine the GUM defined in ~\ref{def:gum}.
	\new{Let $\bSig$ be the covariance matrix of $\Z$. 
		We define the GUM to be \emph{isotropic} when there exists a covariance matrix $\boldsymbol{\Xi}\in\mathbb{R}^{n\times n}$ such that $\bSig = \mathbf{I}_{K-1} \otimes \boldsymbol{\Xi}$, with $\otimes$ the Kronecker product.
		In other words, all $(K-1)$ components of $\Z$ share the same covariance matrix, and are independent from each other.
		Otherwise, the GUM is said to be \emph{anisotropic}.
		We hold the same definition regarding the DGUM.}
\end{definition}

\begin{figure}
	\centering
	\subfloat[]{\includegraphics[width=0.2\linewidth]{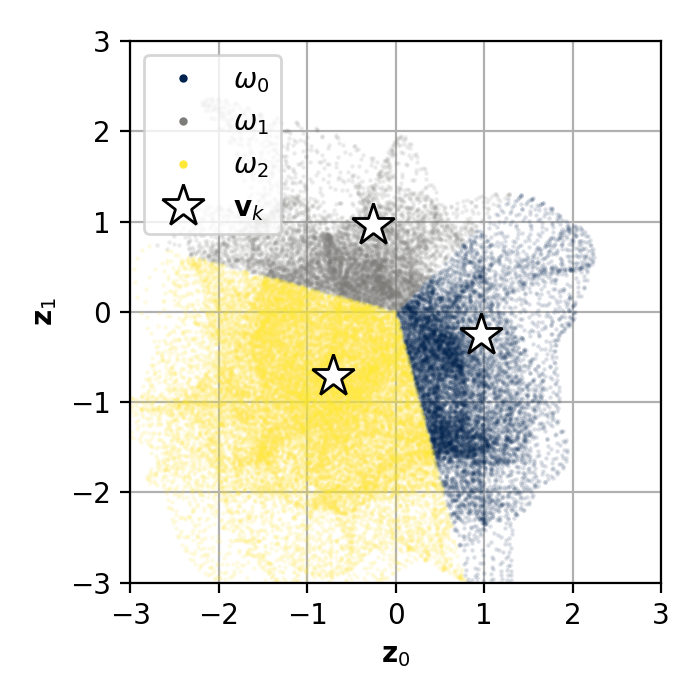}}
	\subfloat[]{\includegraphics[width=0.2\linewidth]{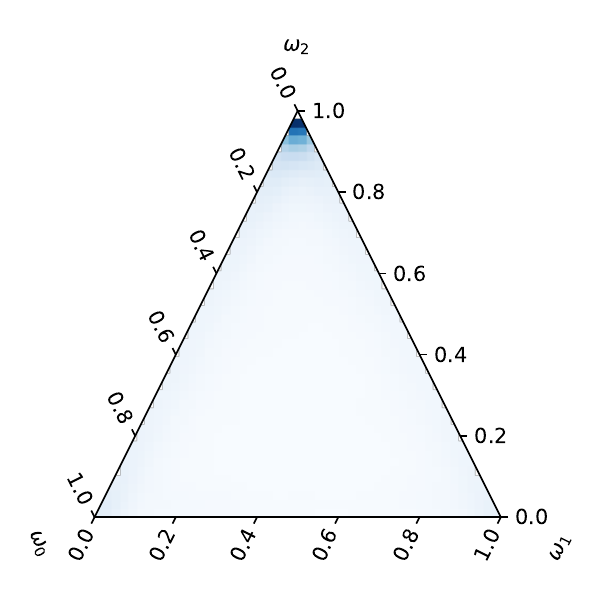}}
	\subfloat[]{\includegraphics[width=0.2\linewidth]{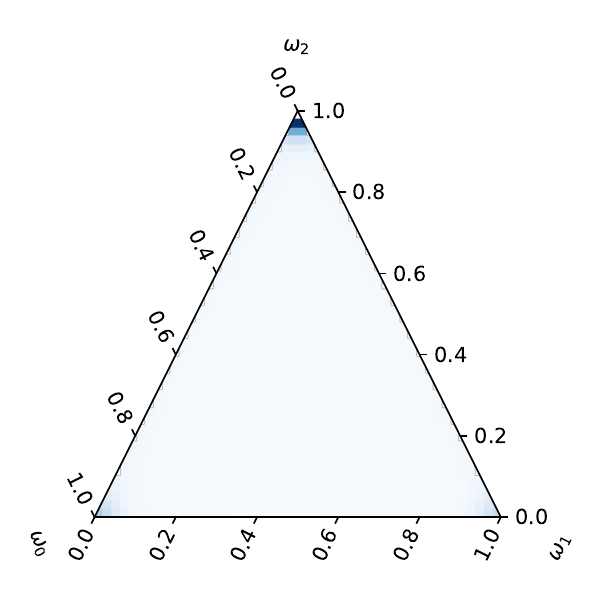}}
	\subfloat[]{\includegraphics[width=0.2\linewidth]{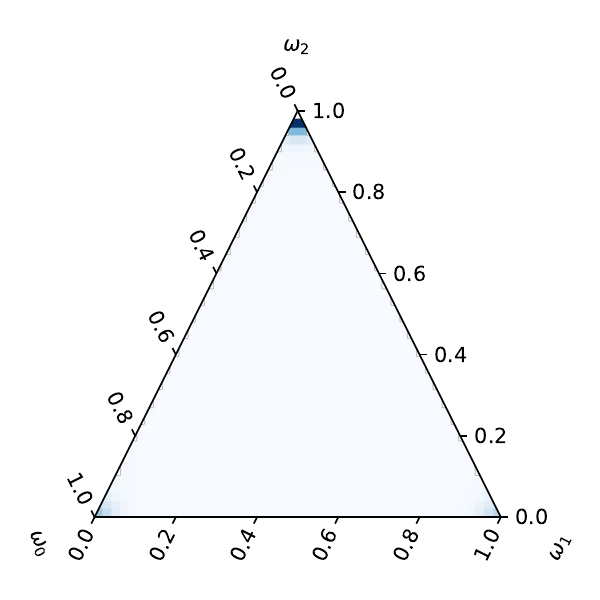}} 
	\subfloat[]{
		\begin{minipage}[b]{0.2\linewidth}\vspace{0pt}
			\includegraphics[width=\linewidth]{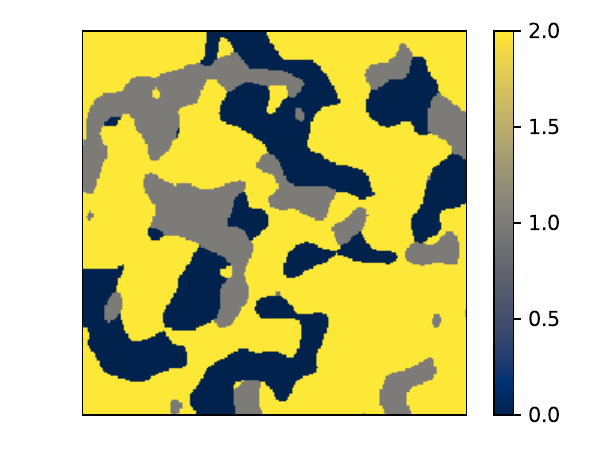}\\[-0.5em]
	\end{minipage}}
	\\
	\subfloat[]{\includegraphics[width=0.2\linewidth]{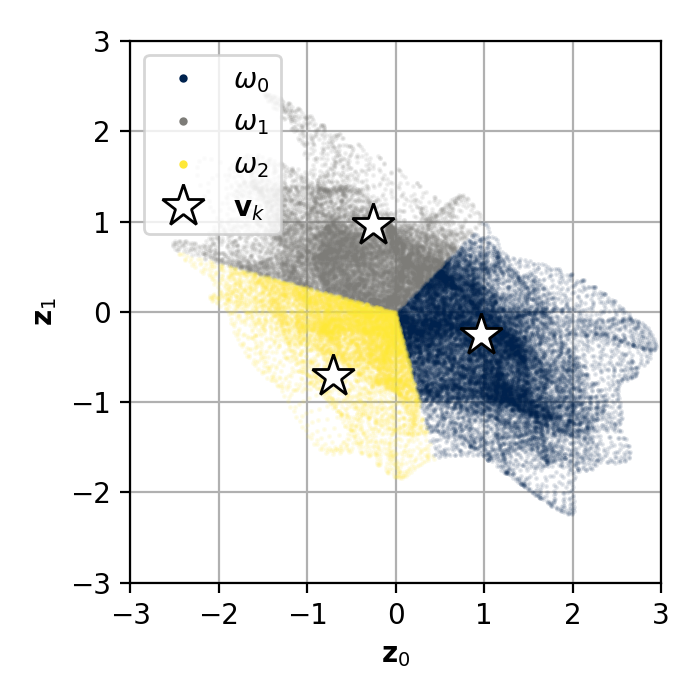}}
	\subfloat[]{\includegraphics[width=0.2\linewidth]{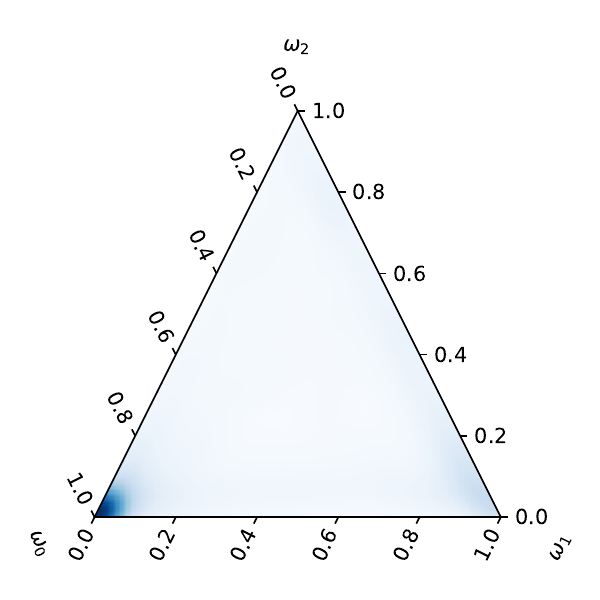}}
	\subfloat[]{\includegraphics[width=0.2\linewidth]{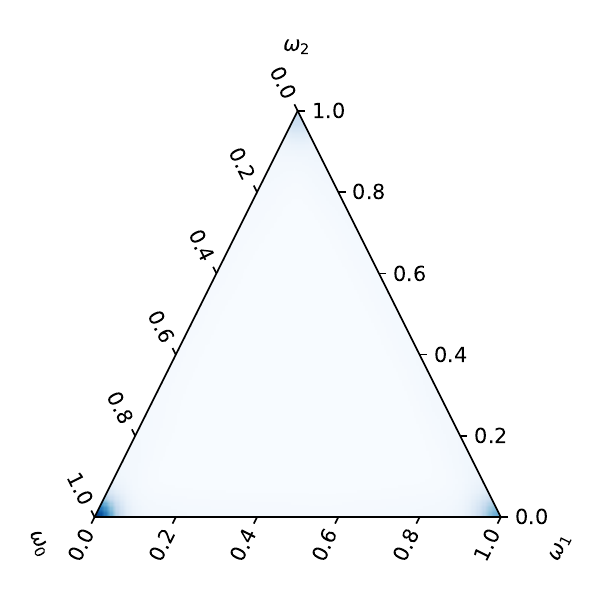}}
	\subfloat[]{\includegraphics[width=0.2\linewidth]{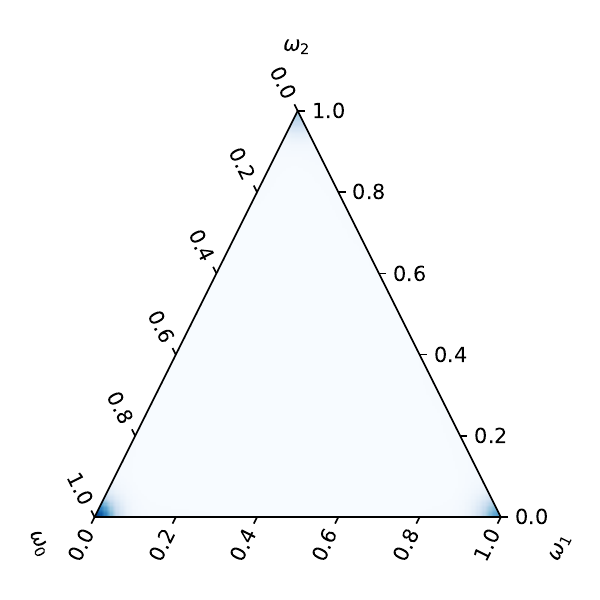}}
	\subfloat[]{
		\begin{minipage}[b]{0.2\linewidth}\vspace{0pt}
			\includegraphics[width=\linewidth]{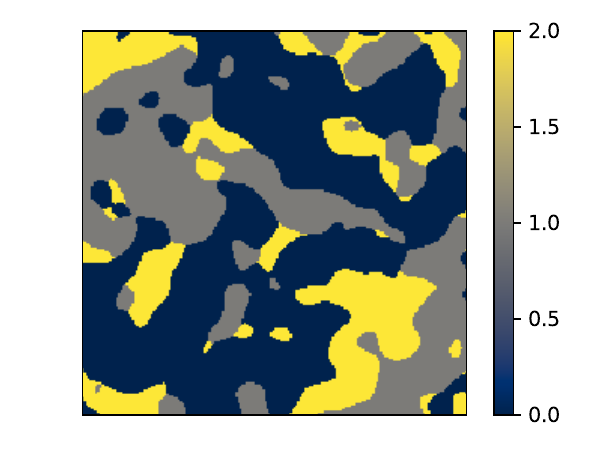}\\[-0.5em]
	\end{minipage}}
	\caption{Depiction of unbalanced (first line) and anisotropic (second line) GUM and DGUM distribution and realization, in the $K=3$ case.
		(a) and (f) depict the marginal locations in $\mathbb{R}^2$ of a realization $\z$, colored according to the class $\omega$, as in Fig.~\ref{fig:2d_gum}.  (b-d) and (g-i) depict the ternary distribution of $\pi$ (as in Fig.~\ref{fig:prob_simplex}), showing in particular how these two situations enable class unbalance. (e) and (j) depict the corresponding DGUM realization.
		%. Their isotropc balanced counterpart is reported in Fig.~\ref{fig:prob_simplex}
	}
	\label{fig:others}
\end{figure}

Figure~\ref{fig:others} depicts the distributions of $\Z$ in the unbalanced and anisotropic cases. In the remainder of the paper, we will focus on balanced isotropic GUMs and DGUMs.

\begin{proposition}[Class balance]\label{prop:balance}
	\new{The balanced isotropic DGUM (definitions~\ref{def:balance} and~\ref{def:isotropic}) reaches class balance: 
		\begin{equation}
			\forall k,~p(X_s=\omega_k) = \frac{1}{K}
	\end{equation}}
\end{proposition}
\begin{proof}
Definition~\ref{def:balance} implies that $\Z$ is centered, and definition~\ref{def:isotropic} that all components within $\Z$ are independent.
	
	Then, the marginal is written, $\forall s$:
	\begin{equation}
		p(\Z_s) \sim \Nc (\mathbf{0}_{K-1}, \mathbf{I}_{K-1} \Xi_{s,s} ).
	\end{equation}
	In other words, when looking at a specific  $s\in \Sc$, $\Z_s$ follows an isotropic, centered normal distribution of variance $\Xi_{s,s}$.
	
	Besides, the unit simplex partitions $\mathbb{R}^{(K-1)}$ into $K$  regions $B_k=\{\z \in \mathbb{R}^{K-1} :  \| \z-\v_k\| \leq  \| \z-\v_i\|, \forall i\}$ that correspond to Voronoï cells of the simplex vertices.
	The Voronoï cells are defined symmetrically, as the $v_k$ are equidistant and centered at the origin. Thus, for any $i,j$ there exists a rotation $\Rc_{ij}$ such that $\Rc_{ij}(v_i) = v_j$ and $\Rc_{ij}(B_i) = B_j$.
	
	As the normal distribution is invariant under orthogonal transformation, $\Rc_{ij}(\Z_s)$ also follows an isotropic, centered normal distribution. Besides, the $B_k$ are rotationally symmetric under permutation of the vertices. Thus, $\forall s$:
	\begin{equation}
		p(\Z_s \in B_i) = p(\Rc_{ij}(\Z_s) \in B_j) = p(\Z_s\in B_j)
	\end{equation}
	As this is true for any $i,j$, then $\forall s$ and $\forall k$:
	\begin{equation}
		p(\Z_s \in B_k) = p(X_s = \omega_k) = \frac{1}{K}
	\end{equation}
\end{proof}

\subsection{Sampling}

In this paper, we propose to use DGUMs as surrogates of classical MRFs. To sample a $K$ class balanced isotropic DGUM, the procedure is the following:
\begin{itemize}
\item Sample $K-1$ independent realizations of centered GMRFs sharing the same covariance matrix $\boldsymbol{\Xi}$, yielding $\Z = \z$.
\item Compute $\X = \x$ through~\eqref{eq:dgum}.
\end{itemize}

Thus, the computational complexity of DGUM sampling is directly that of the GMRF sampling.
Because the latter is performed directly (\textit{i.e.}, no iterative sampling), there is no statistical convergence issue, unilike the Gibbs sampling approach.
As mentioned in Section~\ref{subsec:sampling}, we consider two approaches for sampling, that are appealing because of their tractability.

\textit{Fourier sampling.} Fourier sampling relies on two 2D DFTs over $n$ sites and the sampling of $n$ independent Gaussian random variables (see Appendix~\ref{ap:sample_gmrf}). This yields a computational complexity of $$\Oc\big(n+ 2(K-1)n\log(n)\big) = \Oc\big((K-1)n\log(n)\big).$$

\textit{Spectral sampling.}  This approach~\citep{allard2020simulating} is based on the summation of $p$ cosine waves, each depending on $n$ random variables sampled from a Gaussian variable of inverse-gamma-sampled variance (see Appendix~\ref{ap:sample_gmrf}). The value of $p$ should not be neglected, as it should be large enough to ensure an approximation based on the central limit theorem ($p= 5.10^3$ in~\citep{allard2020simulating}). 
As the sampling along a Gamma distribution is only slightly slower than sampling a normal distribution~\citep{marsaglia2000simple}, we consider its complexity as equivalent. Thus, spectral-based sampling of DGUM has a complexity of
$$\Oc((K-1)np).$$

\section{Numerical results}
\label{sec:numeric}

This section presents numerical experiments and results regarding the sampling of DGUM fields, with comparison to Potts MRF~\eqref{eq:ising_potts} sampling. 

\textit{Experimental setup.}
We evaluate the DGUM sampling using a Matérn covariance function $C$, defined between any sites $s, s' \in \Sc$ as:
\begin{equation}
C(s-s')=\frac{\sigma^2}{2^{\nu-1}\Gamma(\nu)}(\kappa \lVert s -s'\rVert)^\nu K_\nu(\kappa \lVert s-s'\rVert), \label{eq:matern}
\end{equation}
with $K_\nu$ the modified Bessel function of the second kind. 
We set $\sigma=1$, $\kappa=0.1$, and $\nu = 1$ in the numerical experiments. 

As there is no formal equivalence between a Potts MRF and a DGUM, it is not possible to determine an equivalent value for $\beta$ in the Potts MRF~\eqref{eq:ising_potts}. Nevertheless, we found that $\beta=0.5$ yields similar images appearance.
Potts MRF were sampled using the \textit{chromatic} Gibbs sampler~\citep{gonzalez2011parallel}, which parallelizes sampling on mutually independent sets of sites in $\Sc$. The chromatic Gibbs sampler convergence is assessed numerically, and the sampler is stopped when less than 5\% of pixel classes  change between pixel classes of the last image sample and the most frequent pixel class from the previous 10 iterations.

\textit{Implementation.} 
We implemented the DGUM and Potts MRF models and sampling techniques in JAX~\citep{bradbury2021jax}, which is one of the most efficient scientific computing libraries available nowadays. For improved performance, our code makes use of Just-In-Time (JIT) compilation, vectorization and is easily executable on either a GPU or a CPU. The code to reproduce the experiments will be made available at \url{https://github.com/HGangloff/mrfx}. 
%The experiments were made on a laptop Nvidia T600 GPU.

\new{We expect  timings to be highly favorable to the GPU in the spectral sampling and the Gibbs chromatic cases because, respectively, the gamma sampling and the site-wise updates on graph colorings are very well parallelized on a GPU.}
%On the other hand, we expect that timings are favorable to the CPU in the Fourier sampling case for smaller $|\mathcal{S}|$ because the FFT implementation is less well parallelized on the GPU than the CPU.}

\new{
In general, we also expect better JAX performances on a GPU because JIT compilation is optimized for such devices. However, those internal optimizations are in a black-box from the user viewpoint, preventing a fine-grained interpretation of the improvements.} 

\begin{figure}[!ht]
\includegraphics[width=\linewidth]{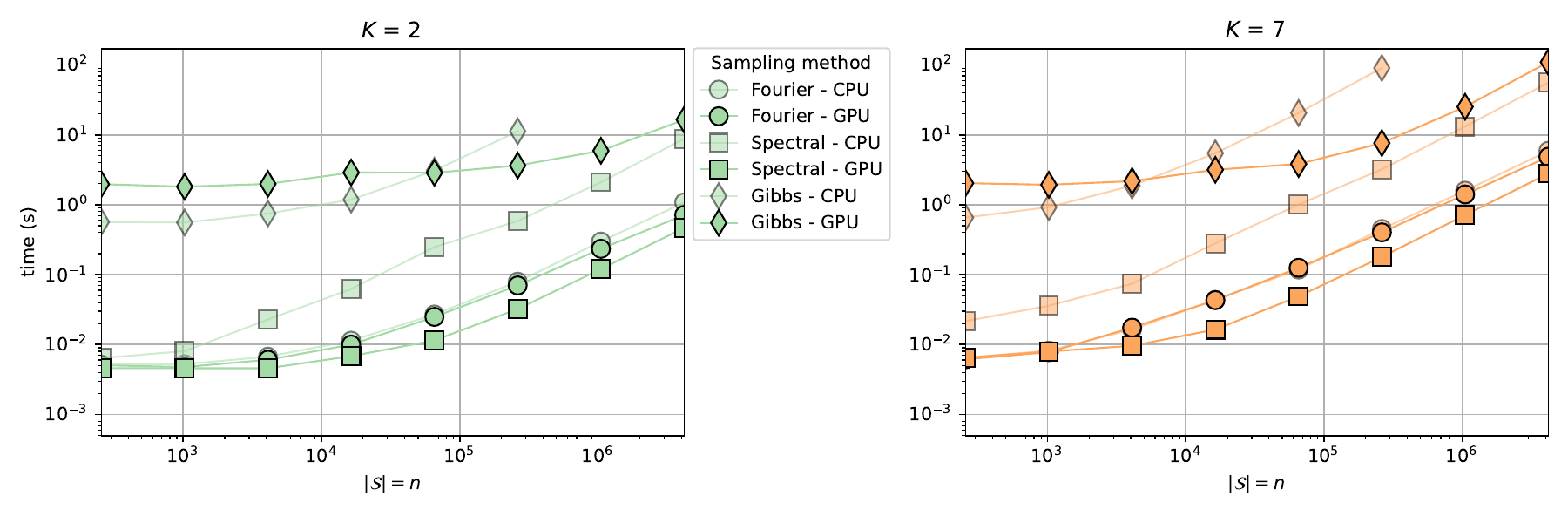}
% \subfloat[DGUM Fourier sampling (100 repetitions).]{ \includegraphics[width=\linewidth]{data_fig_2/times_fourier.pdf}}%
% \\
% \subfloat[DGUM spectral sampling (100 repetitions) and $p=1500$.]{ \includegraphics[width=\linewidth]{data_fig_2/times_spectral.pdf}}%
% \\
% \subfloat[Gibbs chromatic sampling (10 repetitions).]{~\includegraphics[width=\linewidth]{data_fig_2/times_gibbs.pdf}}\hspace{-5em}\\

\caption{Computation times for $K=2$ and $K=7$ classes MRF sampling, as function of the number of sites $n$.  The times reported here use a NVIDIA T600 Laptop GPU and are averaged over 100 repetitions. Additional results, for other values of $K$, are reported in Appendix~\ref{ap:more_time_energy}. 
Due to their time cost, the results regarding Gibbs sampling on CPU are not reported for the largest images ($n = 1024^2$ and $2048^2$). %The larger%\todo{Question : compacter pour $K=2$ et $K=5$ et mettre le reste en annexe ?}
	\label{fig:time}}
	\end{figure}
	
	\subsection{Computational cost}
	\label{subsec:numres_1}
\textit{Time complexity.}
At first, we evaluate the sampling time complexity as a function of the number of sites $n$ and classes $K$. Results are reported in Fig.~\ref{fig:time} and suggest the following observations: 
	\begin{itemize}
\item DGUM sampling is by far faster than Gibbs sampling, even in the case of the chromatic version. 
%For instance, the  gain factor in computation time is comprised between 14 and 68 for the GPU implementation, and even reaches 200 to 290 for the CPU implementation.%  in the CPU implementation, and when comparing  / DGUMs
\item Moving from CPU to GPU mainly benefits to the spectral sampling method, as Fourier sampling benefits from already well-implemented routines on CPU.
\item Fourier sampling yields the best performances in the CPU implementation, as it benefits from already-implemented parallelized routines. On the other hand, the spectral DGUM sampling benefits the most from the GPU version.
\end{itemize}
\new{Thus, the fastest MRF sampling technique is, for most cases, the spectral sampling computed on GPU.}

\begin{figure}[!ht]
\includegraphics[width=\linewidth]{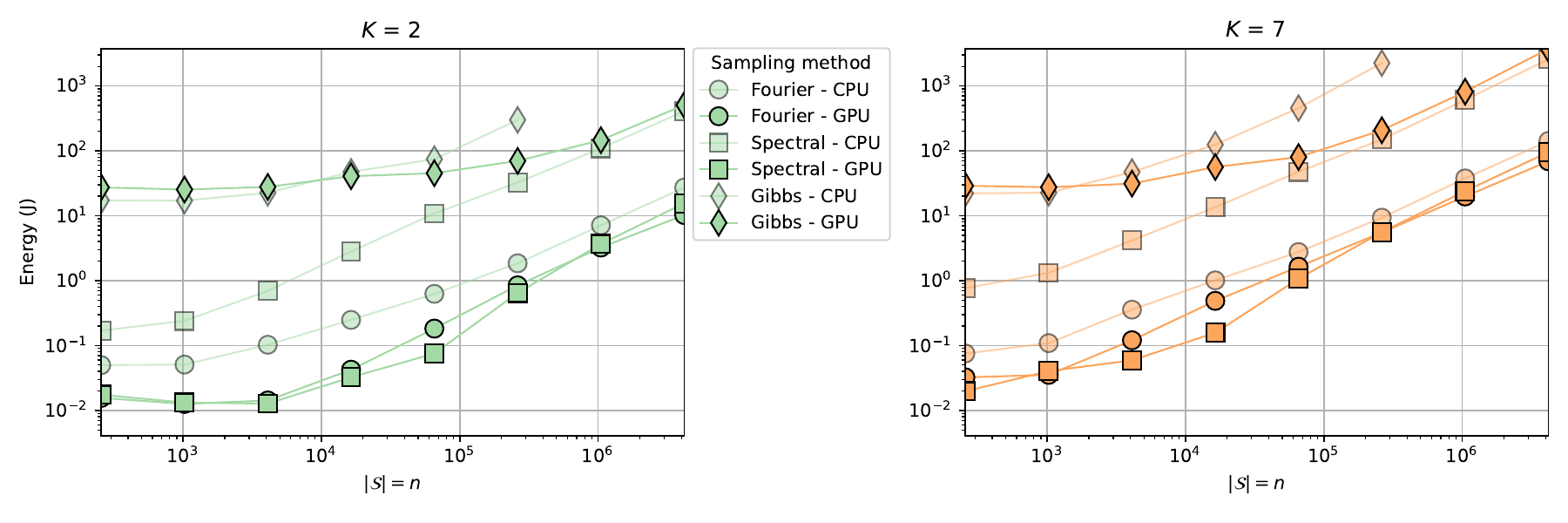}
% \subfloat[DGUM Fourier sampling (100 repetitions).]{ \includegraphics[width=\linewidth]{data_fig_2/times_fourier.pdf}}%
% \\
% \subfloat[DGUM spectral sampling (100 repetitions) and $p=1500$.]{ \includegraphics[width=\linewidth]{data_fig_2/times_spectral.pdf}}%
% \\
% \subfloat[Gibbs chromatic sampling (10 repetitions).]{~\includegraphics[width=\linewidth]{data_fig_2/times_gibbs.pdf}}\hspace{-5em}\\

\caption{Estimated energy consumption of the considered sampling methods, for $K=2$ and $K=7$, as a function of the number of sites $n$. 
Additional results, for other values of $K$, are reported in Appendix~\ref{ap:more_time_energy}.
%Computation energy consumption using CPU-based computation (left) and GPU computation (right) ;  depending on the number of class $K$ and the number of pixels.  \todo{Question : compacter pour $K=2$ et $K=5$ et mettre le reste en annexe ?} 
\label{fig:energy}}
\end{figure}

\new{\textit{Energy consumption.}
We also report the energy consumption for the sampling of an image, with the same varying parameters as in the previous section. We perform the measurement using the Python library Zeus~\citep{zeus-nsdi23} which provides easy-to-use energy measurement of CPUs and GPUs in PyTorch and JAX by wrapping the low-level Nvidia Management Library. 
%\url{https://developer.nvidia.com/management-library-nvml}.
The results are reported in Fig.~\ref{fig:energy}. 
Overall, the observed energy consumption follows expectations, as it increases with both $n$ and $K$.
Besides, GPU-based methods are for almost every $n$ and $K$ more energy-efficient than their CPU counterpart.}
%We also note the higher consumption of GPU-based methods with respect to their CPU counterparts. From the energy consumption perspective, the most energy-efficient sampling technique is consistently (for any $n$ and $K$) the Fourier sampling performed on a CPU.

\begin{figure}[h]
\centering
\subfloat[Improvements of time (left) and energy consumption (right), depicted as a ratio between the measures from the best Gibbs sampler vs. the best DGUM sampler. \label{fig:improvement}]{%
\includegraphics[width=\linewidth]{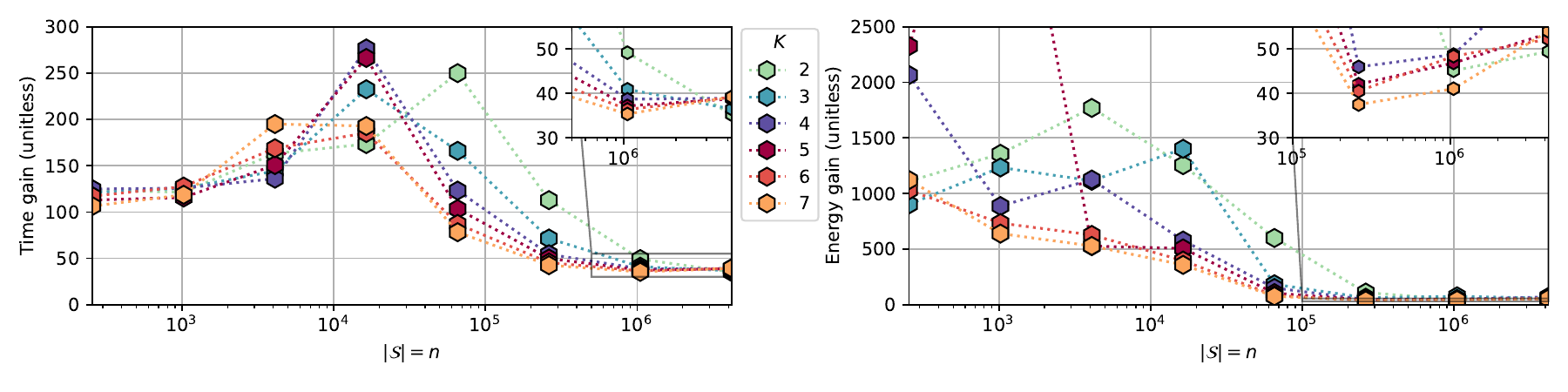}}

\subfloat[Tradeoff between energy consumption and computational speed for the sampling of medium-sized (256\textsuperscript{2} pixels, left) and large (2048\textsuperscript{2} pixels, right) images. The colors depict the same classes as in (a). \label{fig:tradeoff}]{
\includegraphics[width=\linewidth]{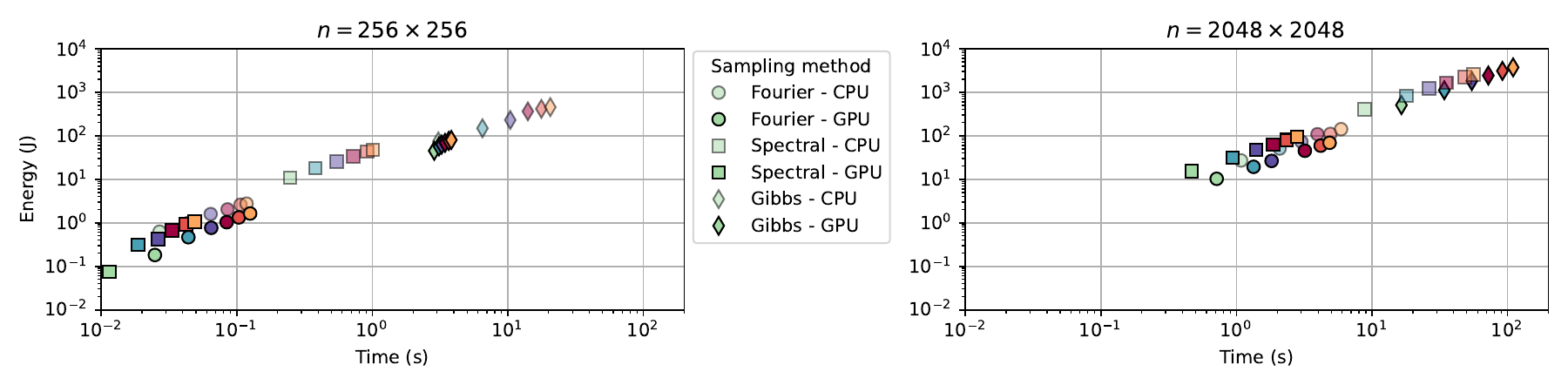}}

\caption{Synthesis of the numerical results in terms of improvements (a) and time-energy tradeoff (b). }
\end{figure}

\new{Overall, the DGUM enables a drastic gain both in computation time and consumption, as shown in Fig.~\ref{fig:improvement}: for any image size $n$, DGUM are sampled at least 35x faster and using at least 37x less energy than the best Gibbs alternative, which is the key finding of this study.
For large images, a tradeoff remains to be made between sampling speed (spectral sampling, on GPU) and energy efficiency (Fourier sampling, on GPU), as is depicted in Fig.~\ref{fig:tradeoff} for two fixed image sizes.}

\begin{remark}
While analyzing the results, the reader should bear in mind that the Zeus energy measurement might not be totally accurate (in absolute value) and that computing power is not the only source of energy consumption, resource utilization, waste, \textit{etc.}~\citep{kaack2022aligning}. Another critical aspect should be noted in the comparison between GPUs and CPUs. While a GPU implementation of CPU code can indeed induce both faster executions and reduced energy consumption at runtime, this should not hide the environmental footprint of manufacturing GPUs~\citep{morand2025does}. The authors then want to argue that investing in GPUs must remain a careful choice.
\end{remark}

%Si on veut donner des ordres de grandeur il est d'usage de convertir en gCO2eq et de comparer à un trajet en voiture comme expliqué page 8 de la publi de Green Algo~\citep{lannelongue2021green}.

\subsection{Statistical properties}

% \begin{table}
% %\renewcommand{\arraystretch}{1.1}
% \renewcommand{\arraystretch}{1.3}
% \setlength{\tabcolsep}{0.5em}
% \centering
% \begin{tabular}{r|c|cccccc}
% 		&K & 2 & 3 & 4 & 5 &6&7 \\ \hline 
% 		\multirow{2}{*}{DGUM / Fourier}      & $\hat{\pi}_0$ & 0.4924 & 0.3278& 0.2369 &0.2095 & 0.1624 & 0.1387\\
% 		&  $\text{std}(\hat{\pi}_0)$ 
% 		%&$6.500e^{-2}$ &  $4.971e^{-2}$ &$5.326e^{-2}$ &      $3.754e^{-2}$ 
% 		&$0.0650$ &  $0.0497$ &$0.0533$ &      $0.0375$  & 0.0454 & 0.0490
% 		\\ \hline
% 		\hline 
% 		\multirow{2}{*}{DGUM / Spectral}      &   $\hat{\pi}_0$ & 0.5036 & 0.3296 & 0.2534 & 0.1995 & 0.1734 & 0.1458\\
% 		&   $\text{std}(\hat{\pi}_0)$ & 0.0935 & 0.0780 & 0.0711 & 0.0481 &0.0541&0.0541\\ \hline
% 		\hline 
% 		\multirow{2}{*}{Gibbs sampling}      &$\hat{\pi}_0$ &    0.4980 & 0.3470 & 0.2480 & 0.1969 & 0.1720 & 0.1439 \\
% 		&    $\text{std}(\hat{\pi}_0)$ & 0.0399 & 0.0581 & 0.0505 & 0.0504 & 0.0482 & 0.0490\\ \hline
% 	\end{tabular}

% \caption{Denoting $\pi_k = p(x_s=k)$, we focus on the $k=0$ case (with similar results for $k>0$). 
% 	% peut être supprimer cette phrase
% 	We use a Matérn covariance with $\kappa=0.1$ for the DGUM sampling, and $\beta = 1.0$ for Gibbs sampling.
% 	Expectations are estimated over 50 repetitions.
% 	\todo{note the gap between estimation and theory rather than the absolute value}
% }
% \label{tab:first}
% \end{table}

\begin{table}
\renewcommand{\arraystretch}{1.3}
\setlength{\tabcolsep}{0.5em}
\centering
% \begin{tabular}{r|c|cccccc}
% 		&K & 2 & 3 & 4 & 5 &6&7 \\ \hline 
% 		\multirow{2}{*}{DGUM / Fourier}      & $|\hat{f}_0 - f_0|$ & 0.0076 & 0.0055& 0.0131 &0.0095 & 0.0043 & 0.0041\\
% 		&  $\text{std}(\hat{f}_0)$ 
% 		%&$6.500e^{-2}$ &  $4.971e^{-2}$ &$5.326e^{-2}$ &      $3.754e^{-2}$ 
% 		&$0.0650$ &  $0.0497$ &$0.0533$ &      $0.0375$  & 0.0454 & 0.0490
% 		\\ \hline
% 		\hline 
% 		\multirow{2}{*}{DGUM / Spectral}      &   $|\hat{f}_0 -f_0|$ & 0.0036 & 0.0037 & 0.0034 & 0.0005 & 0.0067 & 0.0029\\
% 		&   $\text{std}(\hat{f}_0)$ & 0.0935 & 0.0780 & 0.0711 & 0.0481 &0.0541&0.0541\\ \hline
% 		\hline 
% 		\multirow{2}{*}{Gibbs sampling}      &$|\hat{f}_0 - f_0|$ &    0.0020 & 0.0137 & 0.0020 & 0.0031 & 0.0053 & 0.0010 \\
% 		&    $\text{std}(\hat{f}_0)$ & 0.0399 & 0.0581 & 0.0505 & 0.0504 & 0.0482 & 0.0490\\ \hline
% 	\end{tabular}
\begin{tabular}{r|c|cccccc}
		&K & 2 & 3 & 4 & 5 &6&7 \\ \hline 
		\multirow{3}{*}{ $|\hat{f}_0 - f_0|$}&{DGUM / Fourier}       & 0.0076 & 0.0055& 0.0131 &0.0095 & 0.0043 & 0.0041
		\\ \cline{2-8}
        &{DGUM / Spectral}       & 0.0036 & 0.0037 & 0.0034 & 0.0005 & 0.0067 & 0.0029\\
		 \cline{2-8}
		
		&{Gibbs sampling}       &    0.0020 & 0.0137 & 0.0020 & 0.0031 & 0.0053 & 0.0010 
 \\ \hline \hline 
        \multirow{3}{*}{$\text{std}(\hat{f}_0)$ }&{DGUM / Fourier}   &$0.0650$ &  $0.0497$ &$0.0533$ &      $0.0375$  & 0.0454 & 0.0490\\
		\cline{2-8}
		&{DGUM / Spectral}  & 0.0935 & 0.0780 & 0.0711 & 0.0481 &0.0541&0.0541\\
		\cline{2-8}
		&{Gibbs sampling}     & 0.0399 & 0.0581 & 0.0505 & 0.0504 & 0.0482 & 0.0490\\ \hline
	\end{tabular}
\caption{Class balance numerical study. Focusing on the estimator $\hat{f}_0 = \hat{p}(x_s=\omega_0)$, the table depicts the bias $|\hat{f}_0 - f_0| = |\hat{f}_0 - \frac{1}{K}|$ and standard deviation of the frequency estimator.). 
	% peut être supprimer cette phrase
	We use a Matérn covariance with $\kappa=0.1$ for the DGUM sampling, and $\beta = 1.0$ for Gibbs sampling.
	Expectations are estimated over 50 repetitions on images of size $150\times 150$.
}
\label{tab:first}
\end{table}

This section reports statistical measures obtained from sampled DGUM fields in comparison with Potts MRF sampled through chromatic Gibbs sampling.
We focus on estimations of class proportion, pairwise similarity function, and  additional observations regarding a phase transition behavior are included in Appendix~\ref{ap:phase}.

\textit{First order statistic.} Focusing on class proportion, we have shown in Proposition~\ref{prop:balance} that sampling isotropic, balanced GUMS yields class balance. 
Table~\ref{tab:first} reports the statistics of the class proportion estimator, showing the adequation of numerical implementation to the expected $1/K$ ratio. 
Besides, the standard deviation of the estimator (unknown theoretical value) is also reported. This table shows thus that, in terms of class balance estimator, the DGUM sampling behaves similarly to the chromatic Gibbs sampler of Potts MRFs. % both 

\begin{figure}[t]
\centering
% \subfloat[2-class GUM properties]{
	% \includegraphics[width=0.5\linewidth]{proba_dist_k2.pdf}}
% \subfloat[2-class MRF properties]{
	% \includegraphics[width=0.5\linewidth]{proba_dist_MRF2.pdf}}
\includegraphics[width=\linewidth]{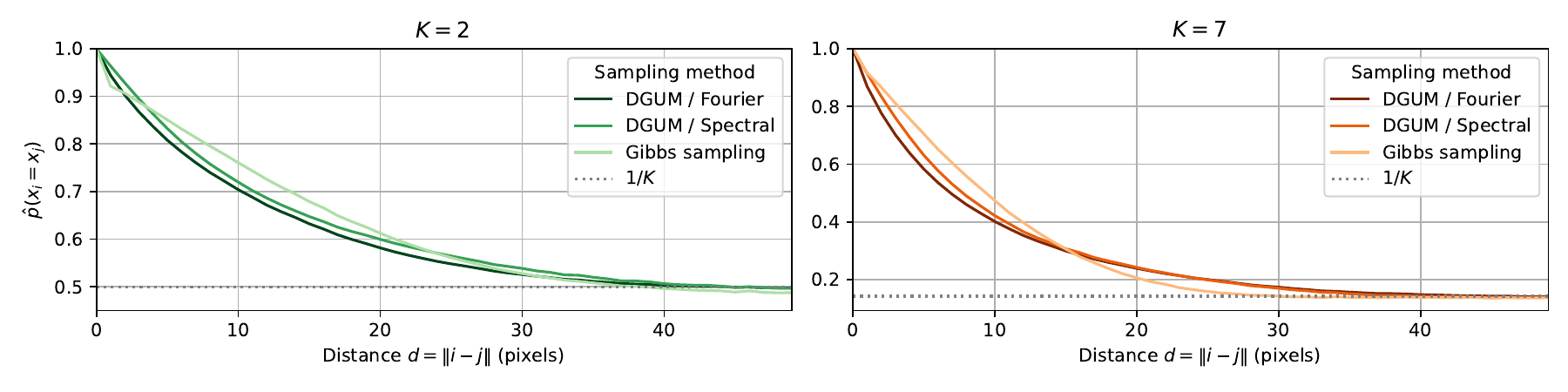}
\caption{Empirical MRF / DGUM comparison setting $\kappa = 0.1$. The Potts MRF was sampled using $\beta = 0.5$ for the $K=2$ case (left) and $\beta=1$ for $K=7$ (right). }
\label{fig:proba_dist_mrf}
\end{figure}

\textit{Second order statistics.}
Figure~\ref{fig:proba_dist_mrf} provides an insight into the pairwise similarity between sites of the realization as a function of their distance $d$ on $\Sc$, namely $p\big(x_i = x_j | \|i - j\|_2 = d\big)$. This function is expected to decrease towards $1/K$: indeed, large distances will yield mutually independent realizations $\z_s$ and $x_s$.
The figure exhibits that the overall behavior of a Potts MRF is similar to that of a DGUM, regardless of the sampling method. 
Remaining differences can be attributed to different constructions of neighborhoods, which spans only 8 sites in Potts MRFs but is larger (depending on the covariance function) with DGUMs.% \todo{Compléter avec les nouveaux graphes}

Wrapping up the results of this section, we can summarize our main numerical results as:
\begin{enumerate}
\item the DGUM sampling outperforms traditional Gibbs sampling in terms of both speed and energy consumption by several orders of magnitude, as highlighted in Fig.~\ref{fig:improvement}.% \todo{préciser en graphe}
\item the resulting fields still share the main numerical properties expected from an Ising / Potts MRF, namely, class balance and pairwise similarity between sites.
\end{enumerate}

\section{Discussion and conclusion}
\label{sec:discussion}
\new{This paper allowed us to introduce the DGUM random fields, that makes use of the sampling techniques from GMRFs to sample discrete MRFs. We exhibited some of the DGUM properties, in particular the fact that DGUMs are also Markov random fields. 
The numerical results confirmed the computational gain  of doing so, while the overall structure of a classical MRF is preserved in DGUM samples. 
Several perspectives stem from this work, and we discuss two avenues for future work. }

\new{\textit{Inference.} Gibbs sampling is also a bottleneck in inverse problems with latent MRF processes. However, the mapping $\phi_{K,c}$ and its limit when $c\rightarrow 0$, noted $\phi_K$, are not surjective. Thus, going back from $\X = \x$ to $\Z=\z$ is not directly feasible, which hinders the use of $\x = \phi_K(\z)$ as a latent variable in an inverse problem. Indeed, the posterior in such a model (\textit{e.g.} with an observation $\Y=\y$ resulting from a noisy measurement of $\x$) cannot be computed directly for a given $\x$ but only for a given $\z$ ; and as $\z \in \mathbb{R}^{N(K-1)}$ this makes the problem difficult to invert. Possible solutions involve the use of approximate sampling techniques, such as variational inference, whose computational cost might counteract the benefit of the DGUM sampling.}%Possibilities involve the use of approximate sampling, such as variational inference.%  limits  
%allproperties

%\new{\textit{Differentiation of discrete fields.} Another direction for future work lies in the area of differentiation. Indeed, we noted earlier some shared concepts with the Gumbel softmax distribution~\citep{maddison2017concrete,jang2017categorical}, which is used as a continuous relaxation of discrete variables and thus enables gradient propagation. We conjecture that GUM random fields could be used similarly while encoding an additional spatial dependency between neighbor sites.} 

\textit{Beyond the grid.} The processes presented here were assumed to lie on a regularly indexed 2D grid $\Sc$, but this is not a necessary condition for DGUM sampling: indeed, one could easily extend this work to handle 3D lattices or graphs. We could also consider the process to lie beyond the grid, as GMRFs are commonly defined outside grids to depict real-valued spatial data. Thus, DGUMs could be sampled at an arbitrary precision, yielding leads for potential use in super-resolution problems.

\appendix

\section{Sampling algorithms}
For the sake of completeness, this appendix  gathers the sampling algorithms considered in this paper, using our notations. 
All these algorithms are implemented and available at \url{https://github.com/HGangloff/mrfx}.

%We report first the Gibbs and Chromatic Gibbs sampler, that enable the sampling of Potts MRF.

\subsection{Sampling Markov Random Fields}
\label{ap:sample_mrf}

The main sampler for Markov random fields is the Gibbs sampler~\citep{geman1984stochastic} introduced by Geman and Geman in 1984. It is reported in Alg.~\ref{alg:gibbs}.

An important improvement was provided in the \textit{chromatic} Gibbs sampler proposed in~\citep{gonzalez2011parallel}. The main idea is that mutually independent sites that can be sampled simultaneously, enabling in practice array-wise sampling and even parallelize part of the loop. 
Note the similarity with the \textit{conclique} based Gibbs sampling~\citep{kaplan2020simulating}, which has a more general perspective regarding the support of the random field on a graph. As their principle is essentially identical, we report only the chromatic Gibbs sampler in Alg.~\ref{alg:chromagibbs}.

\begin{algorithm}[h]
\caption{Gibbs Sampler~\citep{geman1984stochastic} 		\label{alg:gibbs} }
\begin{algorithmic}
	\Require Distribution $p(x_s | \x_{N_s})$ and its parameters, set of sites $\Sc$
	\Ensure Sequence $\{ \x^{(0)}, \x^{(1)}, \ldots, \x^{(P)}\}$
	\State Choose an initial value $\x^{(0)}$.
	\While {a convergence criterion is not reached -- iteration $p$}
	\State Set $\x^{(p)}$ to  $\x^{(p-1)}$.
	\For{each site $s\in\Sc$}  \hfill $\triangleright$ this loop can not be parallelized
	\State Sample $x_s^{(p)}$ from $p(x_s|\x_{N_s}^{(p)})$.
	\EndFor
	%\State $p \leftarrow p+1$
	\EndWhile
\end{algorithmic}
\end{algorithm}

%Chromatic Gibbs sampler~\citep{gonzalez2011parallel}

\begin{algorithm}[h]
\caption{Chromatic Gibbs Sampler~\citep{gonzalez2011parallel} 		\label{alg:chromagibbs} }
\begin{algorithmic}
	\Require Distribution $p(x_s | \x_{N_s})$ and its parameters, and subdivisions of $\Sc$ into mutually independent $\Sc_0, \Sc_1, ..., \Sc_j$ with respect to the chosen neighborhood.
	\Ensure Sequence $\{ \x^{(0)}, \x^{(1)}, \ldots, \x^{(P)}\}$
	\State Choose an initial value $\x^{(0)}$.
	\While {a convergence criterion is not reached -- iteration $p$}
	\State Set $\x^{(p)}$ to  $\x^{(p-1)}$.
	\For{each set $\Sc_i$} \hfill $\triangleright$ this loop can not be parallelized.
	\For{each site $s\in\Sc_i$} \hfill $\triangleright$ this loop can be parallelized
	\State Sample $x_s^{(p)}$ from $p(x_s| \x_{N_s}^{(p)})$.
	\EndFor
	\EndFor
	\EndWhile
\end{algorithmic}
\end{algorithm}

\subsection{Sampling Gaussian Markov Random Fields}
\label{ap:sample_gmrf}
We detail here two samplers for GMRFs: Fourier sampling~\citep{rue2005gaussian} and spectral sampling~\citep{allard2020simulating}.
Fourier sampling relies on the hypothesis that the covariance matrix is circulant, thus assuming that the grid $\Sc$ is a torus, \textit{i.e.}, that the top/bottom and left/right borders are joint.
Denoting $\bb_{\bSig}$ the basis of a covariance matrix $\bSig $ and $\bb_{\Q}$ the basis of the related precision matrix $\Q = \bSig^{-1}$, we have the following properties:
\begin{itemize}
\item $\bb_\Q = \idf(1 \oslash \df(\bb_{\bSig} ))$, with $\df$ and $\idf$ the discrete Fourier transform and its inverse, and $\oslash$ denoting element-wise division. 
\item $\forall \z \in \Rn$, $\Q\z =  \idf(\df(\bb_{\Q}) \odot \df(\z)  ) = \z \ast \bb_{\Q}$, with $\odot$ the element-wise product and $\ast$ the convolution operator.
\end{itemize}
Thus, computations involving $\bSig$ or $\Q$ do not need to involve these matrices fully, but only through their bases $\bb_{\bSig}$ and $\bb_{\Q}$.
Then, Fourier sampling builds upon these properties, using a sampling of a complex i.i.d. Gaussian vector and the computation of the eigenvalues of $\bb_{\Q}$. The procedure is reported in Alg.~\ref{alg:fourier}.

\begin{algorithm}[h]
\caption{Fourier sampling of GMRF~\citep{rue2005gaussian}. 		\label{alg:fourier} }
\begin{algorithmic}
	\Require Circulant precision matrix $\Q$ and its basis $\bb_{\Q}$.%Distribution $p(x_s, | \x_{N_s})$ and its parameters, and subdivisions of $\Sc$ into mutually independent $\Sc_0, \Sc_1, ..., \Sc_n$ with respect to the choosen neighborhood.
	\Ensure GMRF sample with zero-mean and precision $\Q$.%Sequence $\{ \x^{(0)}, \x^{(1)}, \ldots, \x^{(P)}\}$
	\State Sample $\u\in \mathbb{C}^n$ such that $\mathrm{Re}(\u)\sim\Nc(\mathbf{0}_n,\I_n)$ and $\mathrm{Im}(\u)\sim\Nc(\mathbf{0}_n,\I_n)$ 
	\State Compute $\L = \sqrt{n} \, \df(\bb_{\Q}))$
	\item Return $\z = \mathrm{Re}\big( (\L\wedge (-\frac{1}{2}))\odot\u\big)$, where $\wedge$ is a pointwise power operator.
	%\State
\end{algorithmic}
\end{algorithm}

% The sampling procedure builds upon $\U\sim\Nc(0,\I)$, a i.i.d. random normal vector, and
% \begin{equation}
%     \Z = \df( \L \odot \U) \text{, with } \L = \df(\bb_{\Q})\wedge (-\frac{1}{2}) \label{eq:fourier_sampling}
% \end{equation}
% Then $\Z \sim \Nc(\mathbf{0},\bSig)$.
We also report the more recent spectral sampling procedure~\citep{allard2020simulating}.
We use it together with a covariance matrix $\bSig$ designed from a  Matérn correlation function~\eqref{eq:matern}. 
It relies on the idea that summing over a large enough number of well-chosen cosine waves, denoted \textit{bands}, one reaches the conditions of the central limit theorem and thus approximates a Gaussian distribution with the desired covariance. The algorithm is reported, using the notations of this paper, in  Alg.~\ref{alg:spectral}.

% The realization obtained through spectral sampling are computed as: \todo{ajuster les notations ici + où est la gamma ?}
% \new{\begin{equation}
	% 	    Z_s = \sqrt{\frac{2}{L}}\sum_{l=1}^L\cos{(\omega_l^ts+\varphi_l)},
	%         \label{eq:spectral}
	% 	\end{equation}
% where $\{(\omega_l,\varphi_l)\}_{l=1}^L$ are i.i.d samples from a spectral density $\mu$ and a uniform distribution on $[0, 2\pi]$, respectively. We can show that $Z$ is a GRF with $0$ mean and stationary covariance function $C$, where $C$ is the covariance function associated to $\mu$ in Bochner's theorem.
% }

\begin{algorithm}[H]
\caption{Spectral sampling using a Matérn correlation function~\citep{allard2020simulating} \label{alg:spectral}}
\begin{algorithmic}
	\Require number of bands $p$, Matérn parameters $\nu$ and $\kappa$
	\Ensure Sample $\z$, a zero-mean GMRF with covariance function $\bSig$ following a Matérn correlation function~\eqref{eq:matern}. 
	\For{$i = 1$ to $p$}
	\State {Sample $g_i \sim \mathcal{G}(\nu, \kappa^2/2)$;}
	\State {Set $\xi_i = 1/(2g_i)$;} %\todo{En fait $\xi_i$ suit une loi inverse gamma ?} -> Oui
	\State Sample $\eta_i \sim \sqrt{2\xi_i} \mathcal{N}(0, \mathbf{I}_n)$;
	\State Sample $u_i \sim \mathcal{U}(0, 2\pi)$;
	\EndFor
	\For{each $s \in \Sc$}
	\State compute $z_s = \sqrt{\tfrac{2}{p}} \sum_{i=1}^{p} \cos(\eta_i^\top s + u_i)$
	\EndFor
\end{algorithmic}
\end{algorithm}

\section{Additional numerical results}

We report here additional results complimenting those presented in Section~\ref{sec:numeric}, regarding time and energy consumption, as well as phase transition observations.

\subsection{Time and energy consumption}
\label{ap:more_time_energy}

\begin{figure}
    \centering
    \subfloat[Computation times.]{\includegraphics[width=\linewidth]{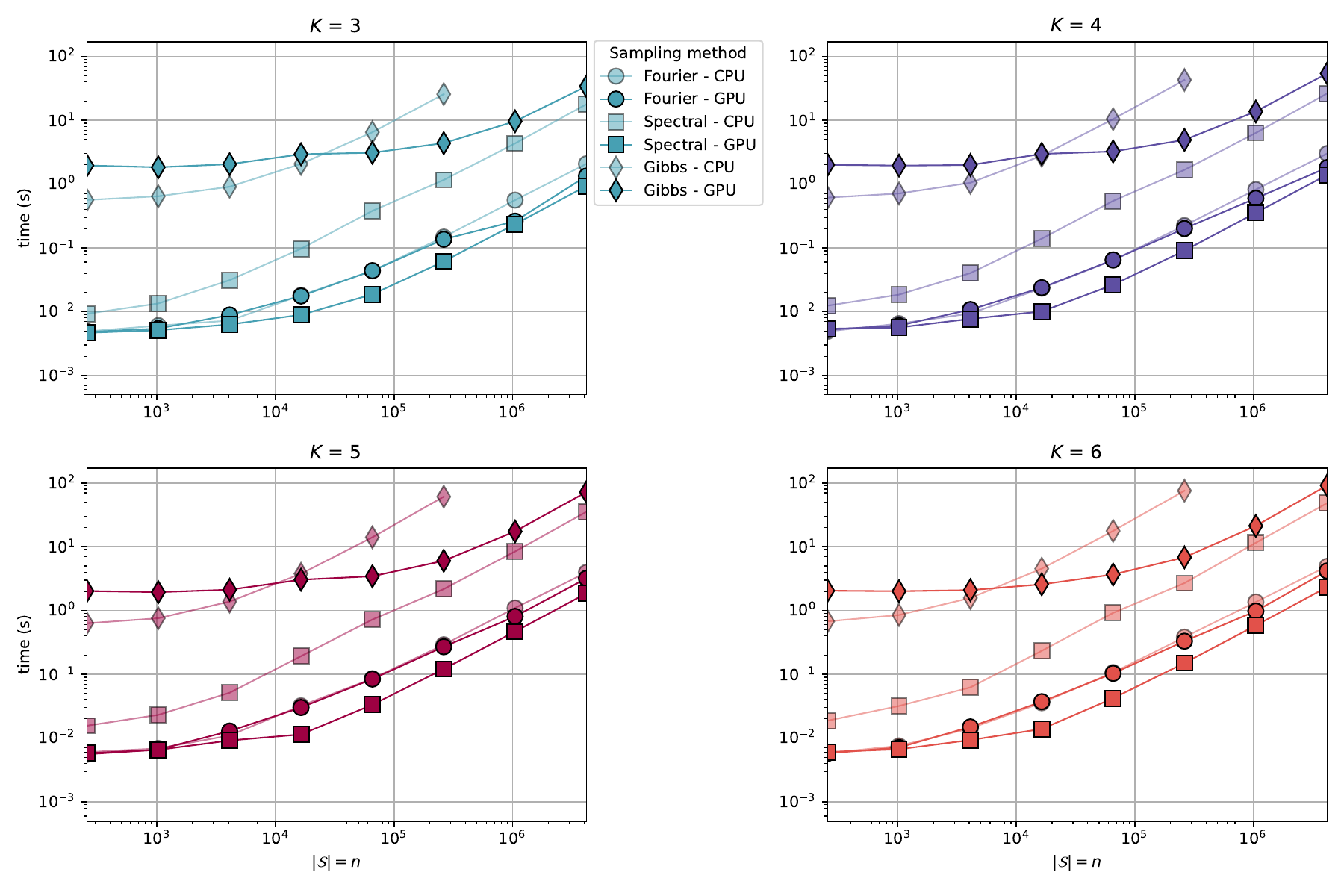}}\\
    \subfloat[Estimated energy consumption.]{\includegraphics[width=\linewidth]{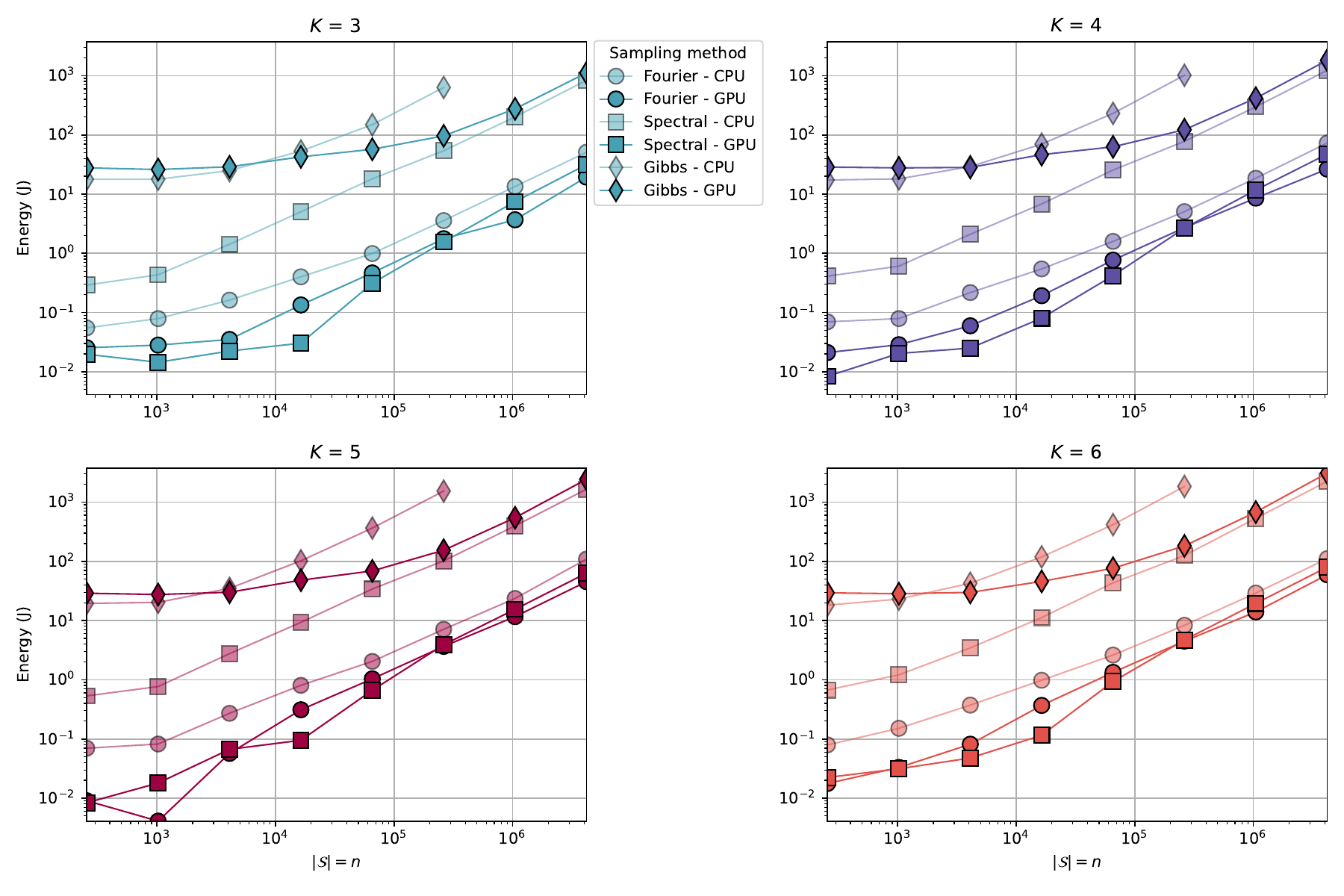}}
    \caption{Computational burden (in time, and energy) extending the results presented in Figs.~\ref{fig:time} and~\ref{fig:energy} to $3 \leq K \leq 6$ classes.}
    \label{fig:an_conso}
\end{figure}

For the sake of brevity, we depicted the results on sampling time and energy consumption for $K=2$ and $K=7$ classes in section~\ref{subsec:numres_1}. 
We provide here additional results for $2<K<7$, depicted in Fig.~\ref{fig:an_conso}.
Overall, the trend is similar along the different values of $K$ and image size $n$: the best results for time and energy consumption are obtained using Spectral sampling on GPU, then Fourier sampling on GPU, then Fourier sampling on CPU.
%computation times are achieved by spectral sampling on GPU, as well as the lower energy consumption.
%while the more energy-efficient sampling is performed by Fourier sampling on CPU. %yields the 

\subsection{Observations of a phase transition}
\label{ap:phase}
\begin{figure}[t]
\,\hfill
\subfloat[Phase transition when sampling a binary field from $\pi_k^c$. ]{\includegraphics[width=0.45\linewidth]{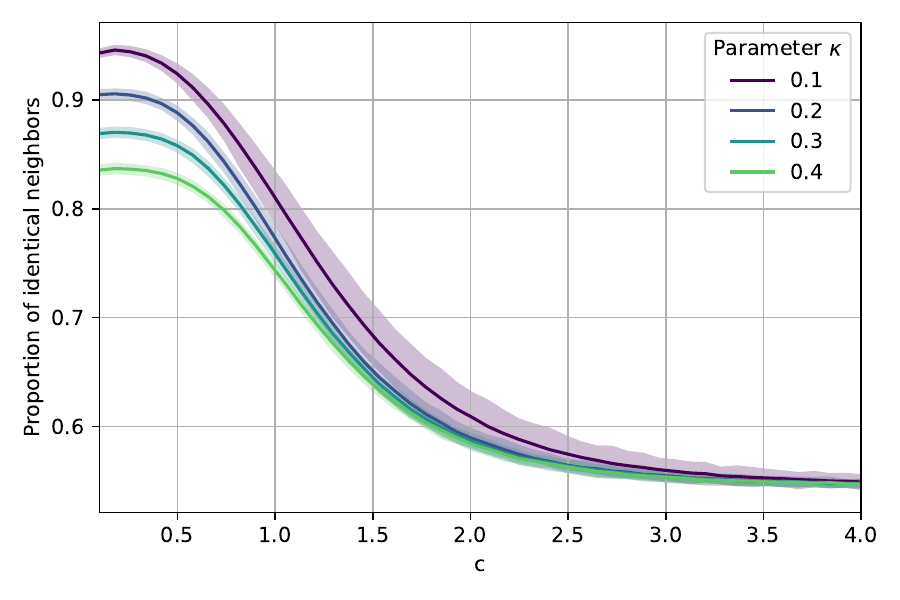}}\hfill 
\subfloat[Phase transition when sampling a DGUM, depending on the parameter $\kappa$ and the number of classes $K$.]{\includegraphics[width=0.45\linewidth]{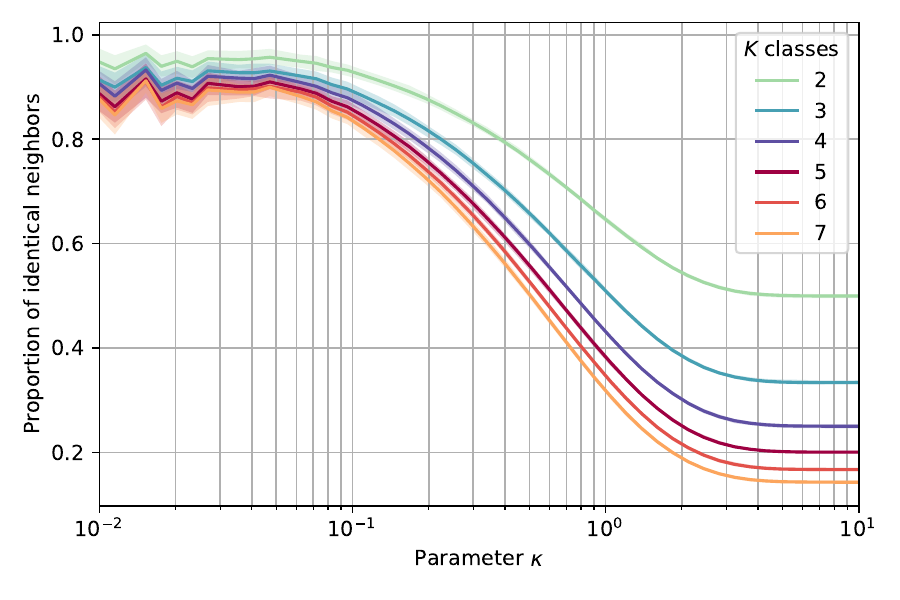}} \hfill\,

\,\hfill 
\begin{minipage}{0.4\linewidth}\,\hfill
	\subfloat[$c=0.1$]{\includegraphics[width=0.3\linewidth,frame]{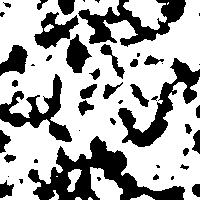}}\hfill
	\subfloat[$c=2$]{\includegraphics[width=0.3\linewidth,frame]{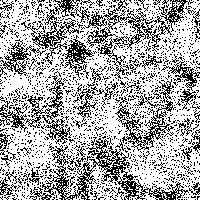}}\hfill
	\subfloat[$c=4$]{\includegraphics[width=0.3\linewidth,frame]{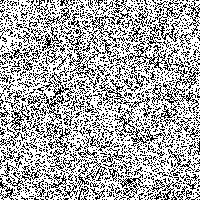}}
\end{minipage}
\hfill 
\begin{minipage}{0.4\linewidth}\,\hfill 
	\subfloat[$\kappa=0.1$]{\includegraphics[width=0.3\linewidth,frame]{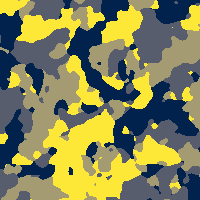}}\hfill
	\subfloat[$\kappa=1$]{\includegraphics[width=0.3\linewidth,frame]{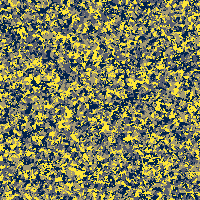}}\hfill
	\subfloat[$\kappa=10$]{\includegraphics[width=0.3\linewidth,frame]{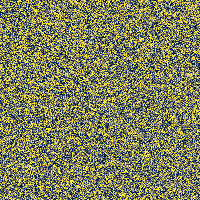}}
\end{minipage}
\hfill \,

\caption{Phase transition phenomena observed within the DGUM sampling process. (a) and (b) depict the overall behavior, with plain lines averaging over 50 samples of size $150\times150$, and shaded regions corresponding to the 1\textsuperscript{st} and 9\textsuperscript{th} decile.  (c)-(e) and (f)-(h) provide examples of realization corresponding to (a) and (b) respectively for a few parameters.}
\label{fig:phase_transition}
\end{figure}

\new{
Depending on their $\beta$ parameter, Ising and Potts MRF exhibit a well known phase transition effect that separates low values of $\beta$, yielding a noisy field, and higher values, exhibiting a constant behavior, with patch-like patterns in between.
A relevant metric indicating phase transition is the proportion of sites identical to their neighbors, and we measure this quantity for two kinds of random fields.
\begin{enumerate}
	\item By construction, the GUM random vectors are built upon the quantity $\pi_{k}^c(\z)$, which measures the distance between any point in $\mathbb{R}^{K-1}$ to the $k-$th vertex of the unit simplex. As  $c\rightarrow 0$, the $\pi_{k}^c(\z)$ will take values close to either 0 or 1; thus, a field $\tilde{\x}$ sampled from $\pi_{k}^c(\z)$ is expected to follow a phase transition depending on $c$.
	%The binary field resulting from the sampling along , depending on $c$. We have shown that $c$ being lower implies sharper transitions, to the limit case 
	\item More directly, the DGUMs are driven by the covariance structure of the GMRF $\Z$ that is sampled beforehand. Thus, we also study the behavior of DGUMs obtained through $\Z=\z$ realizations using a Matérn correlation function~\eqref{eq:matern} within the covariance matrix, while varying $\kappa$ values.
\end{enumerate}
Figure~\ref{fig:phase_transition} depicts the results of the phase transition study.  We can observe, on the $\pi_k^c$, that $c$ increases the randomness of the resulting sampled field, while a stationary behavior happens for $c < 0.5$.  Regarding the phase transition of the DGUM, we note that a large $\kappa$ yields an i.i.d. random fields (with $1/K$ probability of reaching each class), while  $\kappa < 0.1$ yields a patch-like behavior.}

  \bibliography{biblio.bib}

\end{document}